\documentclass[conference]{IEEEtran}
\IEEEoverridecommandlockouts
\usepackage{amsmath,amssymb,amsfonts}
\usepackage{graphicx}
\usepackage{textcomp}
\usepackage{xcolor}
\def\BibTeX{{\rm B\kern-.05em{\sc i\kern-.025em b}\kern-.08em
    T\kern-.1667em\lower.7ex\hbox{E}\kern-.125emX}}

\usepackage{comment}

\usepackage{mathtools} 
\usepackage{booktabs} 
\usepackage{tikz} 

\usepackage[utf8]{inputenc} 
\usepackage{hyperref}       
\usepackage{url}            
\usepackage{amsfonts}       
\usepackage{nicefrac}       
\usepackage{microtype}      
\usepackage{tabularx}
\usepackage{subcaption}
\usepackage{hyperref}
\usepackage{ragged2e}
\usepackage[noend]{algpseudocode}
\usepackage{algorithm}
\usepackage{multicol}

\usepackage[capitalise,nameinlink]{cleveref}

\usepackage{amssymb}
\usepackage{amsfonts}
\usepackage{mathrsfs}
\usepackage{xspace}
\usepackage{bm}
\usepackage{upgreek}

\newcommand{\safemath}[2]{\newcommand{#1}{\ensuremath{#2}\xspace}}

\safemath{\bma}{\mathbf{a}}
\safemath{\bmb}{\mathbf{b}}
\safemath{\bmc}{\mathbf{c}}
\safemath{\bmd}{\mathbf{d}}
\safemath{\bme}{\mathbf{e}}
\safemath{\bmf}{\mathbf{f}}
\safemath{\bmg}{\mathbf{g}}
\safemath{\bmh}{\mathbf{h}}
\safemath{\bmi}{\mathbf{i}}
\safemath{\bmj}{\mathbf{j}}
\safemath{\bmk}{\mathbf{k}}
\safemath{\bml}{\mathbf{l}}
\safemath{\bmm}{\mathbf{m}}
\safemath{\bmn}{\mathbf{n}}
\safemath{\bmo}{\mathbf{o}}
\safemath{\bmp}{\mathbf{p}}
\safemath{\bmq}{\mathbf{q}}
\safemath{\bmr}{\mathbf{r}}
\safemath{\bms}{\mathbf{s}}
\safemath{\bmt}{\mathbf{t}}
\safemath{\bmu}{\mathbf{u}}
\safemath{\bmv}{\mathbf{v}}
\safemath{\bmw}{\mathbf{w}}
\safemath{\bmx}{\mathbf{x}}
\safemath{\bmy}{\mathbf{y}}
\safemath{\bmz}{\mathbf{z}}
\safemath{\bmzero}{\mathbf{0}}
\safemath{\bmone}{\mathbf{1}}

\bmdefine{\biad}{a}
\bmdefine{\bibd}{b}
\bmdefine{\bicd}{c}
\bmdefine{\bidd}{d}
\bmdefine{\bied}{e}
\bmdefine{\bifd}{f}
\bmdefine{\bigd}{g}
\bmdefine{\bihd}{h}
\bmdefine{\biid}{i}
\bmdefine{\bijd}{j}
\bmdefine{\bikd}{k}
\bmdefine{\bild}{l}
\bmdefine{\bimd}{m}
\bmdefine{\bind}{n}
\bmdefine{\biod}{o}
\bmdefine{\bipd}{p}
\bmdefine{\biqd}{q}
\bmdefine{\bird}{r}
\bmdefine{\bisd}{s}
\bmdefine{\bitd}{t}
\bmdefine{\biud}{u}
\bmdefine{\bivd}{v}
\bmdefine{\biwd}{w}
\bmdefine{\bixd}{x}
\bmdefine{\biyd}{y}
\bmdefine{\bizd}{z}

\bmdefine{\bixid}{\xi}
\bmdefine{\bilambdad}{\lambda}
\bmdefine{\bimud}{\mu}
\bmdefine{\bithetad}{\theta}
\bmdefine{\biphid}{\phi}
\bmdefine{\bideltad}{\delta}

\safemath{\bmia}{\biad}
\safemath{\bmib}{\bibd}
\safemath{\bmic}{\bicd}
\safemath{\bmid}{\bidd}
\safemath{\bmie}{\bied}
\safemath{\bmif}{\bifd}
\safemath{\bmig}{\bigd}
\safemath{\bmih}{\bihd}
\safemath{\bmii}{\biid}
\safemath{\bmij}{\bijd}
\safemath{\bmik}{\bikd}
\safemath{\bmil}{\bild}
\safemath{\bmim}{\bimd}
\safemath{\bmin}{\bind}
\safemath{\bmio}{\biod}
\safemath{\bmip}{\bipd}
\safemath{\bmiq}{\biqd}
\safemath{\bmir}{\bird}
\safemath{\bmis}{\bisd}
\safemath{\bmit}{\bitd}
\safemath{\bmiu}{\biud}
\safemath{\bmiv}{\bivd}
\safemath{\bmiw}{\biwd}
\safemath{\bmix}{\bixd}
\safemath{\bmiy}{\biyd}
\safemath{\bmiz}{\bizd}

\safemath{\bmxi}{\bixid}
\safemath{\bmlambda}{\bilambdad}
\safemath{\bmmu}{\bimud}
\safemath{\bmtheta}{\bithetad}
\safemath{\bmphi}{\biphid}
\safemath{\bmdelta}{\bideltad}

\safemath{\bA}{\mathbf{A}}
\safemath{\bB}{\mathbf{B}}
\safemath{\bC}{\mathbf{C}}
\safemath{\bD}{\mathbf{D}}
\safemath{\bE}{\mathbf{E}}
\safemath{\bF}{\mathbf{F}}
\safemath{\bG}{\mathbf{G}}
\safemath{\bH}{\mathbf{H}}
\safemath{\bI}{\mathbf{I}}
\safemath{\bJ}{\mathbf{J}}
\safemath{\bK}{\mathbf{K}}
\safemath{\bL}{\mathbf{L}}
\safemath{\bM}{\mathbf{M}}
\safemath{\bN}{\mathbf{N}}
\safemath{\bO}{\mathbf{O}}
\safemath{\bP}{\mathbf{P}}
\safemath{\bQ}{\mathbf{Q}}
\safemath{\bR}{\mathbf{R}}
\safemath{\bS}{\mathbf{S}}
\safemath{\bT}{\mathbf{T}}
\safemath{\bU}{\mathbf{U}}
\safemath{\bV}{\mathbf{V}}
\safemath{\bW}{\mathbf{W}}
\safemath{\bX}{\mathbf{X}}
\safemath{\bY}{\mathbf{Y}}
\safemath{\bZ}{\mathbf{Z}}

\safemath{\bZero}{\mathbf{0}}
\safemath{\bOne}{\mathbf{1}}
\safemath{\bDelta}{\mathbf{\Delta}}
\safemath{\bLambda}{\boldsymbol\Lambda}
\safemath{\bPhi}{\mathbf{\Upphi}}
\safemath{\bSigma}{\mathbf{\Upsigma}}
\safemath{\bOmega}{\mathbf{\Upomega}}
\safemath{\bTheta}{\mathbf{\Uptheta}}

\bmdefine{\biAd}{A}
\bmdefine{\biBd}{B}
\bmdefine{\biCd}{C}
\bmdefine{\biDd}{D}
\bmdefine{\biEd}{E}
\bmdefine{\biFd}{F}
\bmdefine{\biGd}{G}
\bmdefine{\biHd}{H}
\bmdefine{\biId}{I}
\bmdefine{\biJd}{J}
\bmdefine{\biKd}{K}
\bmdefine{\biLd}{L}
\bmdefine{\biMd}{M}
\bmdefine{\biOd}{N}
\bmdefine{\biPd}{O}
\bmdefine{\biQd}{P}
\bmdefine{\biRd}{R}
\bmdefine{\biSd}{S}
\bmdefine{\biTd}{T}
\bmdefine{\biUd}{U}
\bmdefine{\biVd}{V}
\bmdefine{\biWd}{W}
\bmdefine{\biXd}{X}
\bmdefine{\biYd}{Y}
\bmdefine{\biZd}{Z}

\bmdefine{\biDelta}{\Delta}
\bmdefine{\biLambda}{\Lambda}
\bmdefine{\biPhi}{\Phi}
\bmdefine{\biSigma}{\Sigma}
\bmdefine{\biOmega}{\Omega}
\bmdefine{\biTheta}{\Theta}

\safemath{\bimA}{\biAd}
\safemath{\bimB}{\biBd}
\safemath{\bimC}{\biCd}
\safemath{\bimD}{\biDd}
\safemath{\bimE}{\biEd}
\safemath{\bimF}{\biFd}
\safemath{\bimG}{\biGd}
\safemath{\bimH}{\biHd}
\safemath{\bimI}{\biId}
\safemath{\bimJ}{\biJd}
\safemath{\bimK}{\biKd}
\safemath{\bimL}{\biLd}
\safemath{\bimM}{\biMd}
\safemath{\bimN}{\biNd}
\safemath{\bimO}{\biOd}
\safemath{\bimP}{\biPd}
\safemath{\bimQ}{\biQd}
\safemath{\bimR}{\biRd}
\safemath{\bimS}{\biSd}
\safemath{\bimT}{\biTd}
\safemath{\bimU}{\biUd}
\safemath{\bimV}{\biVd}
\safemath{\bimW}{\biWd}
\safemath{\bimX}{\biXd}
\safemath{\bimY}{\biYd}
\safemath{\bimZ}{\biZd}

\safemath{\bimDelta}{\biDelta}
\safemath{\bimLambda}{\biLambda}
\safemath{\bimPhi}{\biPhi}
\safemath{\bimSigma}{\biSigma}
\safemath{\bimOmega}{\biOmega}
\safemath{\bimTheta}{\biTheta}

\safemath{\setA}{\mathcal{A}}
\safemath{\setB}{\mathcal{B}}
\safemath{\setC}{\mathcal{C}}
\safemath{\setD}{\mathcal{D}}
\safemath{\setE}{\mathcal{E}}
\safemath{\setF}{\mathcal{F}}
\safemath{\setG}{\mathcal{G}}
\safemath{\setH}{\mathcal{H}}
\safemath{\setI}{\mathcal{I}}
\safemath{\setJ}{\mathcal{J}}
\safemath{\setK}{\mathcal{K}}
\safemath{\setL}{\mathcal{L}}
\safemath{\setM}{\mathcal{M}}
\safemath{\setN}{\mathcal{N}}
\safemath{\setO}{\mathcal{O}}
\safemath{\setP}{\mathcal{P}}
\safemath{\setQ}{\mathcal{Q}}
\safemath{\setR}{\mathcal{R}}
\safemath{\setS}{\mathcal{S}}
\safemath{\setT}{\mathcal{T}}
\safemath{\setU}{\mathcal{U}}
\safemath{\setV}{\mathcal{V}}
\safemath{\setW}{\mathcal{W}}
\safemath{\setX}{\mathcal{X}}
\safemath{\setY}{\mathcal{Y}}
\safemath{\setZ}{\mathcal{Z}}
\safemath{\emptySet}{\varnothing}

\safemath{\colA}{\mathscr{A}}
\safemath{\colB}{\mathscr{B}}
\safemath{\colC}{\mathscr{C}}
\safemath{\colD}{\mathscr{D}}
\safemath{\colE}{\mathscr{E}}
\safemath{\colF}{\mathscr{F}}
\safemath{\colG}{\mathscr{G}}
\safemath{\colH}{\mathscr{H}}
\safemath{\colI}{\mathscr{I}}
\safemath{\colJ}{\mathscr{J}}
\safemath{\colK}{\mathscr{K}}
\safemath{\colL}{\mathscr{L}}
\safemath{\colM}{\mathscr{M}}
\safemath{\colN}{\mathscr{N}}
\safemath{\colO}{\mathscr{O}}
\safemath{\colP}{\mathscr{P}}
\safemath{\colQ}{\mathscr{Q}}
\safemath{\colR}{\mathscr{R}}
\safemath{\colS}{\mathscr{S}}
\safemath{\colT}{\mathscr{T}}
\safemath{\colU}{\mathscr{U}}
\safemath{\colV}{\mathscr{V}}
\safemath{\colW}{\mathscr{W}}
\safemath{\colX}{\mathscr{X}}
\safemath{\colY}{\mathscr{Y}}
\safemath{\colZ}{\mathscr{Z}}

\safemath{\opA}{\mathbb{A}}
\safemath{\opB}{\mathbb{B}}
\safemath{\opC}{\mathbb{C}}
\safemath{\opD}{\mathbb{D}}
\safemath{\opE}{\mathbb{E}}
\safemath{\opF}{\mathbb{F}}
\safemath{\opG}{\mathbb{G}}
\safemath{\opH}{\mathbb{H}}
\safemath{\opI}{\mathbb{I}}
\safemath{\opJ}{\mathbb{J}}
\safemath{\opK}{\mathbb{K}}
\safemath{\opL}{\mathbb{L}}
\safemath{\opM}{\mathbb{M}}
\safemath{\opN}{\mathbb{N}}
\safemath{\opO}{\mathbb{O}}
\safemath{\opP}{\mathbb{P}}
\safemath{\opQ}{\mathbb{Q}}
\safemath{\opR}{\mathbb{R}}
\safemath{\opS}{\mathbb{S}}
\safemath{\opT}{\mathbb{T}}
\safemath{\opU}{\mathbb{U}}
\safemath{\opV}{\mathbb{V}}
\safemath{\opW}{\mathbb{W}}
\safemath{\opX}{\mathbb{X}}
\safemath{\opY}{\mathbb{Y}}
\safemath{\opZ}{\mathbb{Z}}
\safemath{\opZero}{\mathbb{O}}
\safemath{\identityop}{\opI}

\safemath{\veca}{\bma}
\safemath{\vecb}{\bmb}
\safemath{\vecc}{\bmc}
\safemath{\vecd}{\bmd}
\safemath{\vece}{\bme}
\safemath{\vecf}{\bmf}
\safemath{\vecg}{\bmg}
\safemath{\vech}{\bmh}
\safemath{\veci}{\bmi}
\safemath{\vecj}{\bmj}
\safemath{\veck}{\bmk}
\safemath{\vecl}{\bml}
\safemath{\vecm}{\bmm}
\safemath{\vecn}{\bmn}
\safemath{\veco}{\bmo}
\safemath{\vecp}{\bmp}
\safemath{\vecq}{\bmq}
\safemath{\vecr}{\bmr}
\safemath{\vecs}{\bms}
\safemath{\vect}{\bmt}
\safemath{\vecu}{\bmu}
\safemath{\vecv}{\bmv}
\safemath{\vecw}{\bmw}
\safemath{\vecx}{\bmx}
\safemath{\vecy}{\bmy}
\safemath{\vecz}{\bmz}

\safemath{\veczero}{\bmzero}
\safemath{\vecone}{\bmone}
\safemath{\vecxi}{\bmxi}
\safemath{\veclambda}{\bmlambda}
\safemath{\vecmu}{\bmmu}
\safemath{\vectheta}{\bmtheta}
\safemath{\vecphi}{\bmphi}
\safemath{\vecdelta}{\bmdelta}

\safemath{\matA}{\bA}
\safemath{\matB}{\bB}
\safemath{\matC}{\bC}
\safemath{\matD}{\bD}
\safemath{\matE}{\bE}
\safemath{\matF}{\bF}
\safemath{\matG}{\bG}
\safemath{\matH}{\bH}
\safemath{\matI}{\bI}
\safemath{\matJ}{\bJ}
\safemath{\matK}{\bK}
\safemath{\matL}{\bL}
\safemath{\matM}{\bM}
\safemath{\matN}{\bN}
\safemath{\matO}{\bO}
\safemath{\matP}{\bP}
\safemath{\matQ}{\bQ}
\safemath{\matR}{\bR}
\safemath{\matS}{\bS}
\safemath{\matT}{\bT}
\safemath{\matU}{\bU}
\safemath{\matV}{\bV}
\safemath{\matW}{\bW}
\safemath{\matX}{\bX}
\safemath{\matY}{\bY}
\safemath{\matZ}{\bZ}
\safemath{\matzero}{\bmzero}

\safemath{\matDelta}{\bDelta}
\safemath{\matLambda}{\bLambda}
\safemath{\matPhi}{\bPhi}
\safemath{\matSigma}{\bSigma}
\safemath{\matOmega}{\bOmega}
\safemath{\matTheta}{\bTheta}

\safemath{\matidentity}{\matI}
\safemath{\matone}{\matO}

\safemath{\rnda}{A}
\safemath{\rndb}{B}
\safemath{\rndc}{C}
\safemath{\rndd}{D}
\safemath{\rnde}{E}
\safemath{\rndf}{F}
\safemath{\rndg}{G}
\safemath{\rndh}{H}
\safemath{\rndi}{I}
\safemath{\rndj}{J}
\safemath{\rndk}{K}
\safemath{\rndl}{L}
\safemath{\rndm}{M}
\safemath{\rndn}{N}
\safemath{\rndo}{O}
\safemath{\rndp}{P}
\safemath{\rndq}{Q}
\safemath{\rndr}{R}
\safemath{\rnds}{S}
\safemath{\rndt}{T}
\safemath{\rndu}{U}
\safemath{\rndv}{V}
\safemath{\rndw}{W}
\safemath{\rndx}{X}
\safemath{\rndy}{Y}
\safemath{\rndz}{Z}

\safemath{\rveca}{\bimA}
\safemath{\rvecb}{\bimB}
\safemath{\rvecc}{\bimC}
\safemath{\rvecd}{\bimD}
\safemath{\rvece}{\bimE}
\safemath{\rvecf}{\bimF}
\safemath{\rvecg}{\bimG}
\safemath{\rvech}{\bimH}
\safemath{\rveci}{\bimI}
\safemath{\rvecj}{\bimJ}
\safemath{\rveck}{\bimK}
\safemath{\rvecl}{\bimL}
\safemath{\rvecm}{\bimM}
\safemath{\rvecn}{\bimN}
\safemath{\rveco}{\bomO}
\safemath{\rvecp}{\bimP}
\safemath{\rvecq}{\bimQ}
\safemath{\rvecr}{\bimR}
\safemath{\rvecs}{\bimS}
\safemath{\rvect}{\bimT}
\safemath{\rvecu}{\bimU}
\safemath{\rvecv}{\bimV}
\safemath{\rvecw}{\bimW}
\safemath{\rvecx}{\bimX}
\safemath{\rvecy}{\bimY}
\safemath{\rvecz}{\bimZ}

\safemath{\rvecxi}{\bmxi}
\safemath{\rveclambda}{\bmlambda}
\safemath{\rvecmu}{\bmmu}
\safemath{\rvectheta}{\bmtheta}
\safemath{\rvecphi}{\bmphi}

\safemath{\rmatA}{\bimA}
\safemath{\rmatB}{\bimB}
\safemath{\rmatC}{\bimC}
\safemath{\rmatD}{\bimD}
\safemath{\rmatE}{\bimE}
\safemath{\rmatF}{\bimF}
\safemath{\rmatG}{\bimG}
\safemath{\rmatH}{\bimH}
\safemath{\rmatI}{\bimI}
\safemath{\rmatJ}{\bimJ}
\safemath{\rmatK}{\bimK}
\safemath{\rmatL}{\bimL}
\safemath{\rmatM}{\bimM}
\safemath{\rmatN}{\bimN}
\safemath{\rmatO}{\bimO}
\safemath{\rmatP}{\bimP}
\safemath{\rmatQ}{\bimQ}
\safemath{\rmatR}{\bimR}
\safemath{\rmatS}{\bimS}
\safemath{\rmatT}{\bimT}
\safemath{\rmatU}{\bimU}
\safemath{\rmatV}{\bimV}
\safemath{\rmatW}{\bimW}
\safemath{\rmatX}{\bimX}
\safemath{\rmatY}{\bimY}
\safemath{\rmatZ}{\bimZ}

\safemath{\rmatDelta}{\bimDelta}
\safemath{\rmatLambda}{\bimLambda}
\safemath{\rmatPhi}{\bimPhi}
\safemath{\rmatSigma}{\bimSigma}
\safemath{\rmatOmega}{\bimOmega}
\safemath{\rmatTheta}{\bimTheta}

\safemath{\dictab}{[\,\dicta\,\,\dictb\,]}

\safemath{\ysig}{\bmy}
\safemath{\ysighat}{\hat{\ysig}}
\safemath{\ysigdim}{M}
\safemath{\xsig}{\bmx}
\safemath{\xsigdim}{N}
\safemath{\nx}{n_x}
\safemath{\zsig}{\bmz}
\safemath{\zsigdim}{\ysigdim}
\safemath{\rsig}{\bmr}
\safemath{\Adict}{\bA}
\safemath{\Adicttilde}{\widetilde{\Adict}}
\safemath{\Adictdim}{\outputdim\times\xsigdim}
\safemath{\avec}{\bma}
\safemath{\avectilde}{\tilde{\avec}}
\safemath{\Bdict}{\bB}
\safemath{\Bdicttilde}{\widetilde{\Bdict}}
\safemath{\Cdict}{\bC}
\safemath{\cvec}{\bmc}
\safemath{\Ddict}{\bD}
\safemath{\Ddictdim}{\ysigdim\times\xsigdim}
\safemath{\dvec}{\bmd}
\safemath{\Ddicttilde}{\widetilde{\bD}}
\safemath{\Bonb}{\bB}
\safemath{\bvec}{\bmb}
\safemath{\Bonbdim}{\ysigdim\times\ysigdim}
\safemath{\noise}{\bmn}
\safemath{\noisedim}{\ysigim}
\safemath{\err}{\bme}
\safemath{\errdim}{\ysigdim}
\safemath{\errset}{\setE}
\safemath{\nerr}{n_e}
\safemath{\delop}{\bP_\errset}
\safemath{\delopc}{\bP_{{\errset}^c}}

\safemath{\cplxi}{\imath}
\safemath{\cplxj}{\jmath}

\safemath{\dict}{\matD}
\safemath{\inputdim}{N}		
\safemath{\outputdim}{M}		
\safemath{\sparsity}{S}	
\safemath{\inputdimA}{{N_a}}	
\safemath{\inputdimB}{{N_b}}	
\safemath{\elemA}{{n_a}}	
\safemath{\elemB}{{n_b}}	
\safemath{\resA}{\matR_a}	
\safemath{\resB}{\matR_b}	
\safemath{\subD}{\matS} 
\safemath{\subA}{\matS_a} 
\safemath{\subB}{\matS_b} 
\safemath{\dicta}{\matA} 	
\safemath{\dictb}{\matB} 	
\safemath{\hollowS}{H}
\safemath{\hollowA}{H_a}
\safemath{\hollowB}{H_b}
\safemath{\cross}{Z}
\safemath{\coh}{\mu_d}			
\safemath{\coha}{\mu_a}			
\safemath{\cohb}{\mu_b}			
\safemath{\mubs}{\nu}	
\safemath{\cohm}{\mu_m} 
\safemath{\dictset}{\setD}	
\safemath{\dictsetp}{\dictset(\coh,\coha,\cohb)}	
\safemath{\dictsetgen}{\dictset_\text{gen}}
\safemath{\dictsetgenp}{\dictsetgen(\coh)}
\safemath{\dictsetonb}{\dictset_\text{onb}}
\safemath{\dictsetonbp}{\dictsetonb(\coh)}

\safemath{\leftside}{U}
\safemath{\rightsideA}{R_a}
\safemath{\rightsideB}{R_b}

\safemath{\indexS}{\setI_S} 

\safemath{\na}{n_a}			
\safemath{\nb}{n_b}			
\safemath{\coeffa}{p_i}	
\safemath{\coeffb}{q_j}	
\safemath{\seta}{\setP}		
\safemath{\setb}{\setQ}     
\safemath{\setw}{\setW}	
\safemath{\setz}{\setZ}	
\safemath{\cola}{\veca}		
\safemath{\colb}{\vecb}		
\safemath{\cold}{\vecd}		
\safemath{\inputvec}{\vecx} 	
\safemath{\error}{\vece}	
\safemath{\noiseout}{\vecz} 	
\safemath{\inputvecel}{x}
\safemath{\inputveca}{\vecx_a}
\safemath{\inputvecb}{\vecx_b}
\safemath{\outputvec}{\vecy}	
\safemath{\lambdamin}{\lambda_{\mathrm{min}}}

\safemath{\elltwo}{\ell_2}
\safemath{\ellone}{\ell_1}
\safemath{\ellzero}{\ell_0}
\safemath{\ellinf}{\ell_\infty}
\safemath{\ellinftilde}{\ell_{\widetilde\infty}}
\safemath{\licard}{Z(\coh,\coha,\cohb)}
\safemath{\xsol}{\hat{x}}
\safemath{\xbord}{x_b}		
\safemath{\xstat}{x_s}		
\safemath{\xstatLone}{\tilde{x}_s}
\safemath{\order}{\mathcal{O}} 
\safemath{\scales}{\Theta} 
\safemath{\ones}{\mathbf{1}} 
\safemath{\zeroes}{\mathbf{0}} 
\safemath{\thlone}{\kappa(\coh,\cohb)} 
\safemath{\constoneA}{\delta} 
\safemath{\constoneB}{\epsilon} 
\safemath{\nlarge}{L}				   
\safemath{\sumlarge}{S_\nlarge}
\safemath{\maxlarger}{P_\nlarge}	   
\safemath{\Pzero}{\textrm{P0}}	
\safemath{\Pone}{\textrm{P1}}
\safemath{\vecfir}{\vecw}			 
\safemath{\vecsec}{\vecz}
\safemath{\elvecfir}{w}              
\safemath{\elvecsec}{z}				 
\safemath{\nlargefir}{n}
\safemath{\normout}{\gamma}
\safemath{\auxfun}{h}
\safemath{\supp}{\textrm{supp}}

\safemath{\indexa}{\ell}
\safemath{\indexb}{r}
\safemath{\indexc}{i}
\safemath{\indexd}{j}

\safemath{\project}{P}

\newcommand\numberthis{\addtocounter{equation}{1}\tag{\theequation}}

\begin{document}

\title{Cost Aware Asynchronous Multi-Agent\\ Active Search
}

\author{\IEEEauthorblockN{Arundhati Banerjee}
\IEEEauthorblockA{\textit{School of Computer Science} \\
\textit{Carnegie Mellon University}\\
Pittsburgh, U.S.A \\
arundhat@cs.cmu.edu}
\and
\IEEEauthorblockN{Ramina Ghods}
\IEEEauthorblockA{\textit{School of Computer Science} \\
\textit{Carnegie Mellon University}\\
Pittsburgh, U.S.A \\
rghods@andrew.cmu.edu}
\and
\IEEEauthorblockN{Jeff Schneider}
\IEEEauthorblockA{\textit{School of Computer Science} \\
\textit{Carnegie Mellon University}\\
Pittsburgh, U.S.A\\
schneide@cs.cmu.edu}
}

\maketitle

\begin{abstract}
Multi-agent active search requires autonomous agents to choose sensing actions that efficiently locate targets. 
In a realistic setting, agents also must consider the costs that their decisions incur.  Previously proposed active search algorithms simplify the problem by ignoring uncertainty in the agent's environment, using myopic decision making, and/or overlooking costs. 
In this paper, we introduce an online active search algorithm to detect targets in an unknown environment by making adaptive cost-aware decisions regarding the agent's actions. Our algorithm combines principles from Thompson Sampling (for search space exploration and decentralized multi-agent decision making), Monte Carlo Tree Search (for long horizon planning) and pareto-optimal confidence bounds (for multi-objective optimization in an unknown environment) to propose an online lookahead planner that removes all the simplifications. 
We analyze the algorithm's performance in simulation to show its efficacy in cost aware active search.
\end{abstract}

\section{Introduction}
Active search \cite{garnett2011bayesian, garnett2012bayesian} in real world applications like environmental monitoring, or search and rescue, involves autonomous robots (agents) accurately detecting targets by making sequential adaptive data collection decisions while minimizing the usage of resources like energy and time. 
%
%
Previous studies have used various constraints and reductions for resource efficient adaptive search. Such algorithms generally include parameters to trade-off the informativeness of the collected data with the cost of such data collection. One approach to adaptive sensing in robotics is to reduce it to a planning problem assuming full observability of the environment \cite{pvenivcka2019data,kent2020human}. Imposing a cost budget then reduces to constrained path planning between the known start and goal locations. Unfortunately, this is in contrast with the real world where the agent's environment, the number of targets and their locations may be unknown and the agent may have access only to noisy observations from sensing actions. All these factors increase the difficulty of cost effective active search. 

Besides cost efficiency, executing active search with multiple agents creates an additional challenge. Centralized planning in a multi-agent setting is often impractical due to communication constraints \cite{yan2013survey,robin2016multi}. Further, a real world system dependent on a central coordinator that expects synchronicity from all agents is susceptible to communication or agent failure.

In our problem formulation, the agents are not entirely independent actors and therefore still share information with their peers in the team when possible. However, we do not require them to communicate synchronously and instead assume that each agent is able to independently plan and execute its next sensing action using whatever information it already has and happens to receive.  

In this paper, we propose a novel cost-aware asynchronous multi-agent active search algorithm called CAST (Cost Aware Active Search of Sparse Targets) to enable agents to detect sparsely distributed targets in an unknown environment using noisy observations from region sensing actions, without any central control or synchronous inter-agent communication. CAST performs cost-aware active search knowing only the costs of its feasible actions without requiring a pre-specified cost budget. It combines Thompson sampling with Monte Carlo Tree Search for lookahead planning and multi-agent decision making, along with Lower Confidence Bound (LCB) style pareto optimization to tradeoff expected future reward with the associated costs.  We demonstrate the efficacy of CAST with a set of simulation results across different team sizes and number of targets in the search space.

\section{Problem formulation}
\label{sec:problem}

Consider a team of autonomous agents actively sensing regions of the search space looking for targets. To plan its next sensing action, each cost-aware agent has to trade-off the expected future reward of detecting a target with the overall costs it will incur in travelling to the appropriate location and executing the action. Given previous observations, it adaptively makes such data-collection decisions online while minimizing the associated costs as much as possible. Unfortunately, this problem is NP-hard \cite{lim2016adaptive}.

\textbf{Sensing setup:}
We first describe our setup for active search with region sensing. 
Consider a gridded search environment with ground truth described by a sparse matrix 
having $k$ non-zero entries at the locations of the $k$ targets. We define the flattened ground truth vector as $\bm\beta \in \mathbb{R}^n$ where each entry is either 1 (target) or 0 (no target). $\bm\beta$ is the search vector that we want to recover.
The sensing model for an agent $j$ at time $t$ is
\vspace{-2mm}
\begin{equation}\vspace{-2mm}
    y_t^j = {\bmx_t^j}^\mathrm{T}\bm\beta + \epsilon_t^j,\;\text{where } \epsilon_t^j \sim \mathcal{N}(0,\sigma^2)\label{eq:sensing_model}.
\end{equation}
$\bmx_t^j \in \mathbb{R}^n$ is the (flattened) sensing action at time $t$. We define our action space $\mathcal{A}$ ($\bmx_t^j \in \mathcal{A}$) to include only hierarchical spatial pyramid sensing actions \cite{lazebnik2006beyond}.
$y_t^j$ is the agent's observation and $\epsilon_t^j$ is a random, i.i.d added noise. $(\bmx_t^j, y_t^j)$ is agent $j$'s measurement at time $t$. 
Note that the agents using this region sensing model must trade-off between sensing a wider area with lower accuracy versus a highly accurate sensing action over a smaller region. 
The support of the vector $\bmx_t$ is appropriately weighted so that $\|\bmx_t\|_2 = 1$ to ensure each sensing action has a constant power. This helps us in modeling observation noise as a function of the agent's distance from the region \cite{ghods2021decentralized}. Since each action has a constant power and every observation has an i.i.d added noise with a constant variance, the signal to noise ratio in the unit squares comprising the rectangular sensing block reduces as the size of the sensing region is increased. 

\textbf{Cost model:}
We introduce the additional realistic setting that the sensing actions have different associated costs. First, we consider that the agent travelling from location $a$ to location $b$ incurs a travel time cost $c_{d}(a,b)$.
\footnote{We assume a constant travelling speed and compute the Euclidean distance between locations $a$ and $b$.}
Second, we assume that executing each sensing action at location $b$ incurs a time cost $c_{s}(b)$. Therefore, $T$ time steps after starting from location $x_0$, an agent $j$ has executed actions $\{\bmx_t^j\}_{t=1}^{T}$ and incurs a total cost defined by $C^j(T) = \sum_{t=1}^{T} c_{d}(x_{t-1}^j,x_{t}^j) + c_{s}(x_t^j)$. 

\textbf{Communication setup:}
We assume that communication, although unreliable, will be available sometimes and the agents should utilize it when possible. The agents share their own past measurements asynchronously with teammates, but do not wait for communication from their teammates at any time 
since this wait could be arbitrarily long and thus cause a loss of valuable sensing time. In the absence of synchronicity, we also do not require that the set of available past measurements remain consistent across agents since communication problems can disrupt it. 
%
%
We will denote the set of measurements available to an agent $j$ at time $t$ by $\bD_t^j = \{(\bmx_{t'},y_{t'}) | \{t'\} \subseteq \{1,\dots,t-1\}\}$, $|\mathbf{D}_t^j| \leq t-1$, which includes its own past observations and those received from its teammates till time $t$.

\section{Related work}
\label{sec:relatedwork}
Autonomous target search has diverse applications in environment monitoring \cite{popovic2017multiresolution}, wildlife protection \cite{linchant2015unmanned} as well as search and rescue operations \cite{gupta2017decision}. 
In robotics, informative path planning (IPP) problems focus on adaptive decision making to reach a specified goal state or region. In contrast to our setting, common IPP algorithms consider a known environment \cite{meera2019obstacle},
are myopic or use non-adaptive lookahead \cite{singh2009nonmyopic}, and assume weakly coupled sensing and movement actions \cite{choudhury2020adaptive}.

Bayesian optimization and active learning methods are another approach to active search \cite{marchant2014sequential, rajan2015bayesian, jiang2017efficient}. Unfortunately, they mostly address single-agent systems, or if multi-agent they assume central coordination \cite{azimi2012batch,jiang2018efficient} and except for \cite{ma2017active} lack any realistic assumptions on sensing actions. Multi-agent asynchronous active search algorithms proposed in \cite{ghods2021decentralized,ghods2021multi} tackle several of these challenges but they are myopic in nature. Further, \cite{jiang2019cost} introduced cost effective non-myopic active search but their simplified setting 
excludes simultaneous evaluation of multiple search points with different costs.

Our active search formulation has close similarities with planning under uncertainty using a Partially Observable Markov Decision Process (POMDP) \cite{kaelbling1998planning}. Monte Carlo Tree Search (MCTS) \cite{kocsis2006bandit, browne2012survey} has found success as a generic online planning algorithm in large POMDPs \cite{silver2010monte}, but is mostly limited to the single agent setting \cite{flaspohler2019information, fischer2020information}. 

Decentralized POMDP (Dec-POMDP) \cite{bernstein2002complexity, oliehoek2016concise} is another framework for decentralized active information gathering using multiple agents which is typically solved using offline, centralized planning followed by online, decentralized execution \cite{Lauri_JAAMAS2020, lauri2020multi}. 
Decentralized MCTS (Dec-MCTS) algorithms have also been proposed for multi-robot active perception under a cost budget \cite{sukkar2019multi,best2020decentralised} but they typically rely on each agent maintaining a joint probability distribution over its own belief as well as those of the other agents. 

Finally, cost aware active search can be viewed as a multi-objective sequential decision making problem.  \cite{lee2018monte} developed an MCTS algorithm for cost budgeted POMDPs using a scalarized version of the reward-cost trade-off whereas \cite{wang2012multi} introduced multi-objective MCTS (MO-MCTS) for discovering global pareto-optimal decision sequences in the search tree. Unfortunately, MO-MCTS is computationally expensive and unsuitable for online planning. \cite{chen2019pareto} proposed the Pareto MCTS algorithm for multi-objective IPP but they ignore uncertainty due to partial observability in the search space.

\section{Our proposed algorithm: CAST}
\label{sec:algo}
\subsection{Background}
\label{subsec:4.1}
We first briefly describe the concepts essential to the planning and decision making components of our algorithm.

\textbf{Monte Carlo Tree Search (MCTS)} is an online algorithm that combines tree search with random sampling in a domain-independent manner. In our setup, a cost-aware agent would benefit from the ability to lookahead into the adaptive evolution of its belief about the target's distribution in the environment in response to possible observations from the actions it might execute. We therefore consider MCTS as the basis for developing our online planning method with finite horizon lookahead. Unfortunately, the presence of uncertainty about targets (number and location) in the unknown environment together with the noisy observations introduces additional challenges in our problem formulation. 

\textbf{Pareto optimality:} Our formulation of cost aware active search described in \cref{sec:problem} can be viewed as a multi-objective
sequential decision making problem. A common approach to solving such multi-objective optimization problems is scalarization i.e. considering a weighted combination resulting in a single-objective problem that trades off the different objectives. However, tuning the weight attributed to each objective is challenging since they might be scaling quantities having different units and their relative importance might be context dependent. In contrast, pareto optimization builds on the idea that some solutions to the multi-objective optimization problem are categorically worse than others and are \emph{dominated} by a set of pareto-optimal solution vectors forming a pareto front for the optimization objective. Considering a set of $D$-dimensional vectors $\bmg \in \mathcal{G}$, we define the following:
\begin{itemize}
    \item $\bmg$ dominates $\bmg'$ (i.e. $\bmg \succ \bmg'$) iff: (1) $\forall d\in\{1,\dots,D\}$, $[\bmg]_d \geq [\bmg']_d$ (2) $\exists d \in \{1,\dots,D\}$, $[\bmg]_d > [\bmg']_d$
    \item $\bmg$ and $\bmg'$ are incomparable (i.e. $\bmg || \bmg'$) iff: $\exists d_1,d_2\in\{1,\dots,D\}$, $[\bmg]_{d_1}>[\bmg']_{d_1}$ \emph{and} $[\bmg]_{d_2}<[\bmg']_{d_2}$
    \item $\mathcal{G^*}\subseteq\mathcal{G}$ is the pareto-front of $\mathcal{G}$ iff: (1)$\forall \bmg\in\mathcal{G}$ and $\forall \bmg'\in\mathcal{G^*}$, $\bmg \not\succ \bmg'$ (2) $\forall \bmg,\bmg'\in\mathcal{G^*}$, $\bmg||\bmg'$.
\end{itemize}
In our algorithm, each agent estimates a reward-cost vector to evaluate its candidate actions and chooses the next sensing action from a pareto-optimal set of such vectors.

\textbf{Thompson sampling (TS)} \cite{thompson1933likelihood}, 
studied in a number of bandit and reinforcement learning (RL) settings \cite{gopalan2014thompson,russo2017tutorial,gopalan2015thompson}, balances exploration with exploitation by choosing actions that maximize the expected reward assuming that a sample drawn from the posterior distribution is the true state of the world. 
As a result, exploration in TS is reward oriented, leaning heavily on the drawn posterior sample \cite{bai2014thompson, leike2016thompson}. 
Moreover TS only explores on the seemingly optimal policies, which is a disadvantage in environments where knowledge-seeking actions having lower immediate reward are crucial for the agent's efficient long-term performance. This is especially relevant for cost-aware active search wherein actions which are not immediately rewarding in terms of having detected a target may still be informative, for example, by reducing the uncertainty regarding the presence of targets in a certain part of the search space. In this work, we build upon these insights and adapt TS in combination with MCTS to develop a search tree building strategy for online lookahead planning 
under partial observability.

Additionally, posterior sample based action selection makes TS an excellent candidate for a decentralized multi-agent decision making algorithm \cite{kandasamy2018parallelised} and it has been shown to be effective in multi-agent active search with myopic planning \cite{ghods2021decentralized,ghods2021multi}. In this work, we will show that our TS based lookahead planning algorithm enables decentralized multi-agent active search with no pre-coordination and minimum communication overhead among agents, in contrast with existing multi-agent algorithms that rely on pre-designed movement coordination or communication and update of joint probability distributions.

\subsection{Our approach}
\label{subsec:4.2}
Following the setting described in \cref{sec:problem}, consider $J$ agents searching for $k$ sparsely located targets in an unknown environment and $\bm\beta$ is the search vector we want to recover. The agent's belief $b(\bm\beta)$ over $\bm\beta$ is a continuous probability distribution over the search space. For any agent $j$, the prior belief is modeled by $b_0^j = \mathrm{P}(\bm\beta) = \mathcal{N}(\bm\mu_0,\bm\Sigma_0)$ and the likelihood function following the sensing model (\ref{eq:sensing_model}) is given by $\mathrm{P}(y_t^j | \bm\beta,\bmx_t^j) = \mathcal{N}({\bmx_t^j}^T\bm\beta,\sigma^2)$. Therefore, at time step $t$, its posterior belief over $\bm\beta$ is denoted $b_t^j = \mathrm{P}(\bm\beta|\mathbf{D}_{t}^j\cup\{\bmx_t^j,y_t^j\}) = \mathcal{N}(\bm\mu_t^j,\bm\Sigma_t^j)$. 
In the multi-agent setting, each agent $j$ maintains its own posterior belief $b_t^j(\bm\beta)$ and estimate 
$\hat{\bm\beta}(\mathbf{D}_{t}^j\cup\{\bmx_t^j,y_t^j\})
= \big(\sigma^2*{\bm\Sigma_0}^{-1} +  \begin{bmatrix}
{\bX_t^j}^{\text{T}} & {\bmx_t^j}
\end{bmatrix}\begin{bmatrix}
\bX_t^j\\
{\bmx_t^j}^{\text{T}}
\end{bmatrix} \big)^{-1}\begin{bmatrix}
{\bX_t^j}^{\text{T}} & \bmx_t^j
\end{bmatrix}\begin{bmatrix}
\bmy_t^j\\
y_t^j
\end{bmatrix}
$
where $\{\bX_t^j,\bmy_t^j\}$ consist of the measurements in $\mathbf{D}_{t}^j$.

\textbf{Reward formulation:} We first observe that our active search problem can be categorized as parameter estimation in active learning, developed in \cite{kandasamy2019myopic} with the name Myopic Posterior Sampling (MPS). Like MPS, our objective is to actively recover the search vector $\bm\beta$. However, MPS being myopic, 
chooses $\bmx_t^j$ that maximizes $\mathbb{E}_{y_t^j|\bmx_t^j,\bm\beta_t^j}\big[\lambda(\bm\beta_t^j, \mathbf{D}_{t+1}^j)\big]$ where $\lambda(\bm\beta_t^j, \mathbf{D}_{t+1}^j) = -\|\bm\beta_t^j - \hat{\bm\beta}(\mathbf{D}_{t+1}^j)\|_2^2$, $\bm\beta_t^j\sim b_t^j$ and $\bD_{t+1}^j = \bD_t^j\cup\{\bmx_t^j,y_t^j\}$. Essentially, $\lambda(\bm\beta_t^j, \mathbf{D}_{t+1}^j)$ is designed so that the agents will keep exploring the search space as long as there is uncertainty in the posterior samples $\bm\beta_t^j$. Simultaneously, the posterior belief distribution will contain uncertainty as long as there are unexplored or less explored locations in the search space. 

In contrast with MPS, to be cost efficient, our active search agents would benefit from non-myopic
reasoning about the trade-off between potential reward obtained by identifying targets versus cost incurred from executing such sensing actions over a finite horizon lookahead. But we note that  $\lambda(\bm\beta_t^j, \mathbf{D}_{t+1}^j) \leq 0$, therefore if we simply extend the MPS reward over multiple lookahead steps and try to maximize the value of cumulative discounted reward divided by total incurred cost, it would erroneously favour costlier actions for the same reward. Instead, we propose using $\lambda^{-}(\bm\beta_t^j, \mathbf{D}_{t+1}^j) = \max\{0, \|\bm\beta_{t}^j - \hat{\bm\beta}(\mathbf{D}_{t}^j)\|_2^{2} - \|\bm\beta_{t}^j - \hat{\bm\beta}(\mathbf{D}_{t+1}^j)\|_2^{2}\}$ as the one-step lookahead reward. We design $\lambda^{-}$ to encourage information gathering by favoring actions $\bmx_t$ that reduce the uncertainty in the posterior sample $\bm\beta_{t}^j$ over consecutive time steps. 
Additionally, $\lambda^{-}(\bm\beta_t^j, \mathbf{D}_{t+1}^j) \geq 0$. Now, we can compute the $u$-step lookahead reward $R^u(\bmx_t,\bm\beta_t^j)$ over action sequence $\bmx_{t:t+u}$ as the $\gamma$-discounted expected sum of $\lambda^-$ over $u$ steps.
\vspace{-2mm}
\begin{equation}
 R^u(\bmx_t,\bm\beta_t^j) = \mathbb{E}_{y_{t:t+u}}\big[\sum_{\Delta t=1}^u\gamma^{\Delta t-1}\lambda^{-}(\bm\beta_t^j, \mathbf{D}_{t+\Delta t}^j)\big] 
 \numberthis\label{eq:imm_reward}
\end{equation}

Following the discussion in \cref{subsec:4.1} about TS, we observe that the reward computation in \eqref{eq:imm_reward} is dependent on the 
posterior sample $\bm\beta_t^j$. Particularly, $\lambda^{-}(\bm\beta_t^j, \mathbf{D}_{t+1}^j)$ is higher for sensing actions $\bmx_t$ that identify the non-zero support elements of the vector $\bm\beta_t^j$. Further, maximizing $R^u(\bmx_t,\bm\beta_t^j)$ over all sequences $\bmx_{t:t+u}$ for a sampled $\bm\beta_t^j$ would exacerbate this problem by choosing a series of point sensing actions that identify the non-zero support of the particular sample. Section 8.2 of \cite{russo2017tutorial} also highlights this drawback of employing TS based exploration in active learning problems that require a careful assessment of the information gained from actions. 
%
In order to overcome these challenges, we propose generalizing the posterior sampling step to a sample size greater than one and combine the information from these samples using confidence bounds 
over $\lambda^-$ to evaluate the corresponding sensing actions.
To further clarify these design details, we now describe our new algorithm CAST, 
outlined in \cref{alg:search}.

\textbf{CAST:} At each time step $t$, on the basis of its history $\bD_t^j$ of past measurements, the agent $j$ decides its next region sensing action $\bmx_{t}^j$ using the SEARCH procedure of \cref{alg:search}. It starts with an empty tree $\mathcal{T}_t^j$ having just a root node and gradually builds it up over $m$ episodes. We assume a maximum tree depth $d_{\text{max}}$. Our search tree has two types of nodes - belief nodes and action nodes. 
A belief node $h$ is identified by the history of actions and observations accumulated in reaching that node. An action node $(h,a)$ is identified by the action $\bma$ taken at the immediately preceding belief node $h$ in the search tree. The root as well as the leaves are belief nodes. 
\begin{algorithm}[!h]
	\caption{Cost Aware Active Search of Sparse Targets}
		\begin{algorithmic}[1]
			\Procedure{main}{}\Comment{Executed on each agent $j$}
			\For{$t$ in $\{1,2,\hdots\}$}
			\State{$\bmx_t^j$ = SEARCH$(\bD_{t}^j)$}\label{line:5}
			\State{Execute $\bmx_t^j$. Observe $y_t^j$. $\bD_{t+1}^j = \bD_{t}^j\cup\{\bmx_t^j,y_t^j\}$}\label{line:6}
			\State{Share $\{\bmx_t^j,y_t^j\}$ asynchronously with teammates.}\label{line:communication}
			\State{Update belief $b_{t+1}^j$ and estimate $\hat{\bm\beta}(\bD_{t+1}^j)$.}
			\EndFor
			\EndProcedure
			\Procedure{search}{$\bD_t$}
			\State{\textbf{At time t: }History $\bD_t = \{(\bmx_i, y_i)\}_{i=1}^t$, Agent location $x_t$, Belief $b_t = \mathrm{P}(\bm\beta | \bD_t)$, Search tree $\mathcal{T}_t = \phi$ }
			\For {each episode $m' \in \{1 \hdots m\}$}\label{line:10}
            \State{Sample $\bm\beta\sim b_t$. Discretize $\bm\beta$ to get $\bm\beta_{m',t}$.}\label{line:15}
			\State{SIMULATE$(\bm\beta_{m',t},\bD_t,x_t,0)$}\label{line:16}
			\EndFor
			\State{$\setA_t^* = $ParetoOptimalActionSet($\mathcal{T}_t$)}\label{line:20}
			\State{$\bma_t^* = \operatorname*{argmax}_\bma\{\frac{\bma.r^{LCB}}{\bma.cost} | \bma \in \setA_t^*\}$}\label{line:21}
			\State{\textbf{return } $\bma_t^*$}
			\EndProcedure
			\Procedure{simulate}{$\bm\beta,\bD,x,d$}
			\State{\textbf{Input: }Posterior sample $\bm\beta$, Root node history $\bD$, Agent's current location $x$, Root node depth $d$}
			\State{$n(h) \leftarrow n(h)+1$}\label{line:25}\Comment{Denote root (belief) node as $h$}
			\If{$d = d_{\text{max}}$}\label{line:26}
			\Return $0,0$ \Comment{Reached leaf node}
			\EndIf
			\If{$\lfloor n(h)^{\alpha_s}\rfloor > \lfloor (n(h)-1)^{\alpha_s}\rfloor$}\label{line:27}
			\State{add new child action node $(h,a)$}\label{line:28}
			\Else{ select action node $(h,a)$ using \eqref{treepolicynew}}\label{line:31}
			\EndIf
			\State{$n(h,a) \leftarrow n(h,a)+1$}\label{line:32}
			\State{$o \leftarrow \bma^\mathrm{T}\bm\beta$, $\bD' \coloneqq \bD \cup \{\bma,o\}$}\label{line:33}
			\If{$o$ was not previously observed at $(h,a)$}\label{line:37}
			\State{append new node $h'$ due to $o$ in branch $hah'$}\label{line:38}
			\EndIf
			\State{$r_{h'} = \lambda^-(\bm\beta,\bD), c_{h'}(x,a) = c_d(x, a) + c_s(a)$}\label{line:35_36}
			\State{Update $r_{h'}^{LCB}$ and $\bmg_{h'} = \begin{bmatrix}r^{LCB}_{h'}& -c_{h'}(x,a)\end{bmatrix}^\mathrm{T}$}\label{line:update_rLCB}
			\State{$r',c'$ = SIMULATE($\bm\beta,\bD',a,d+1$)}\label{line:39}
			\State{$r'' = r_{h'}+\gamma\times r'$, $c'' = c_{h'}+c'$}\label{line:40}
			\State{$\bar{Q}^{UCT}(h,a) = \frac{\bar{Q}^{UCT}(h,a)\times(n(h,a)-1) + \frac{r''}{c''}}{n(h,a)} $}\label{line:41}
			\State{LCBParetoFrontUpdate($h'$)}\label{line:42}
			\State{LCBParetoFrontUpdate($(h,a)$)}\label{line:43}
			\State{\textbf{return }$r'',c''$}\label{line:44}
			\EndProcedure
		\end{algorithmic}
	\label{alg:search}
\end{algorithm}
Each episode $m'\in\{1,\dots,m\}$ comprises the following:
\begin{enumerate}
	\item \textit{Sampling}: First, a posterior sample is drawn at the root node from the belief $b_t^j = \mathrm{P}(\bm\beta | \mathbf{D}_{t}^j)$ and discretized into a binary vector $\bm\beta_{m',t}^j \in \{0,1\}^n$ (\cref{line:15}).
	\item \textit{Selection and Expansion}: Starting at the root node, a child action node selection policy (tree policy) is applied at every belief node $h$ in a top-down depth-first traversal till a leaf node is reached. 
	In order to prevent tree width explosion with increasing size of the action space, the progressive widening parameter $\alpha_s$ (\cref{line:27}) \cite{coulom2007computing} determines when a new action node is added to the tree. 
    Arriving at any action node $(h,a)$, the corresponding maximum likelihood (ML) observation $o = \bma^T\bm\beta_{m',t}^j$ is computed (\cref{line:33}) which helps transition to its child belief node $h'$. The 1-step reward $\lambda^{-}_{m'}$ for $\bm\beta=\bm\beta_{m',t}^j$ and associated execution cost is computed at each belief node visited in $m'$ (\cref{line:35_36}). Every belief and action node in $m'$ also updates the number of times it has been visited so far (\cref{line:25,line:32}).

%
	\item \textit{Backpropagation}: Once the maximum depth is reached, the lookahead rewards and associated costs are backpropagated up from the leaf to 
	each belief and action node visited in $m'$. 
	 Each action node stores the discounted reward per unit cost averaged over $n(h,a)$ simulations in the subtree rooted at that node (\cref{line:41}). Further, each belief and action node builds a reward-cost pareto front (\cref{line:42,line:43}) using the backed up values from their respective subtrees which is utilized in deciding $\bmx_{t}$ after $m$ episodes (\cref{line:20}).
\end{enumerate}
 \textbf{Remark 1.} \textit{The size of action space in active search is larger than what MCTS algorithms commonly deal with, unless they are augmented with a neural policy network \cite{silver2016mastering,silver2017mastering}. Having a continuous state vector gives rise to additional challenges of exploding width at the belief nodes, making the tree too shallow to be useful and may cause collapse of belief representations resulting in overconfidence in the estimated policy. The added observation noise would exacerbate these challenges. Therefore, discretization of the posterior sample $\bm\beta$ and using the ML observation in updating the belief nodes are important modifications to make MCTS work in our setting.}

%
%
%
\textbf{Tree policy:} UCT (Upper Confidence Bound applied to trees) \cite{kocsis2006bandit} is the tree policy used in most MCTS implementations to balance exploration-exploitation in building the search tree. UCT exploits action nodes based on their lookahead reward estimates averaged over past episodes but does not account for the inter-episode variance in such rewards. Particularly in our setting, the lookahead reward at any action node in an episode $m'$ depends on the posterior sample $\bm\beta_{m',t}^j$ drawn at the root node and this stochasticity leads to sample variance especially when the particular action node has been visited in only a few episodes. We can account for this variance using the UCB-tuned policy \cite{audibert2006use} to guide action node selection. Besides, \cite{shah2020non} formalized a correction to the UCT formula in an MDP framework replacing its logarithmic exploration term with an appropriate polynomial. We extend it to our tree policy in CAST, called CAST-UCT \eqref{treepolicynew}, by combining it with UCB-tuned to balance exploration with exploitation while building the search tree in our partially observable state space. Specifically, CAST-UCT chooses
\begin{equation}
\bma^* = \mathop{argmax}_\bma Q(h,a) + \sqrt{\frac{2\sigma_{h,a}\sqrt{n(h)}}{n(h,a)}} + \frac{16\sqrt{n(h)}}{3n(h,a)}.
\numberthis\label{treepolicynew}
\end{equation}
$\sigma_{h,a}^2$ is the variance of the 
terms averaged in $Q(h,a)$.
%
%

\begin{figure}[htp]
	\centering
    \includegraphics[width=\linewidth]{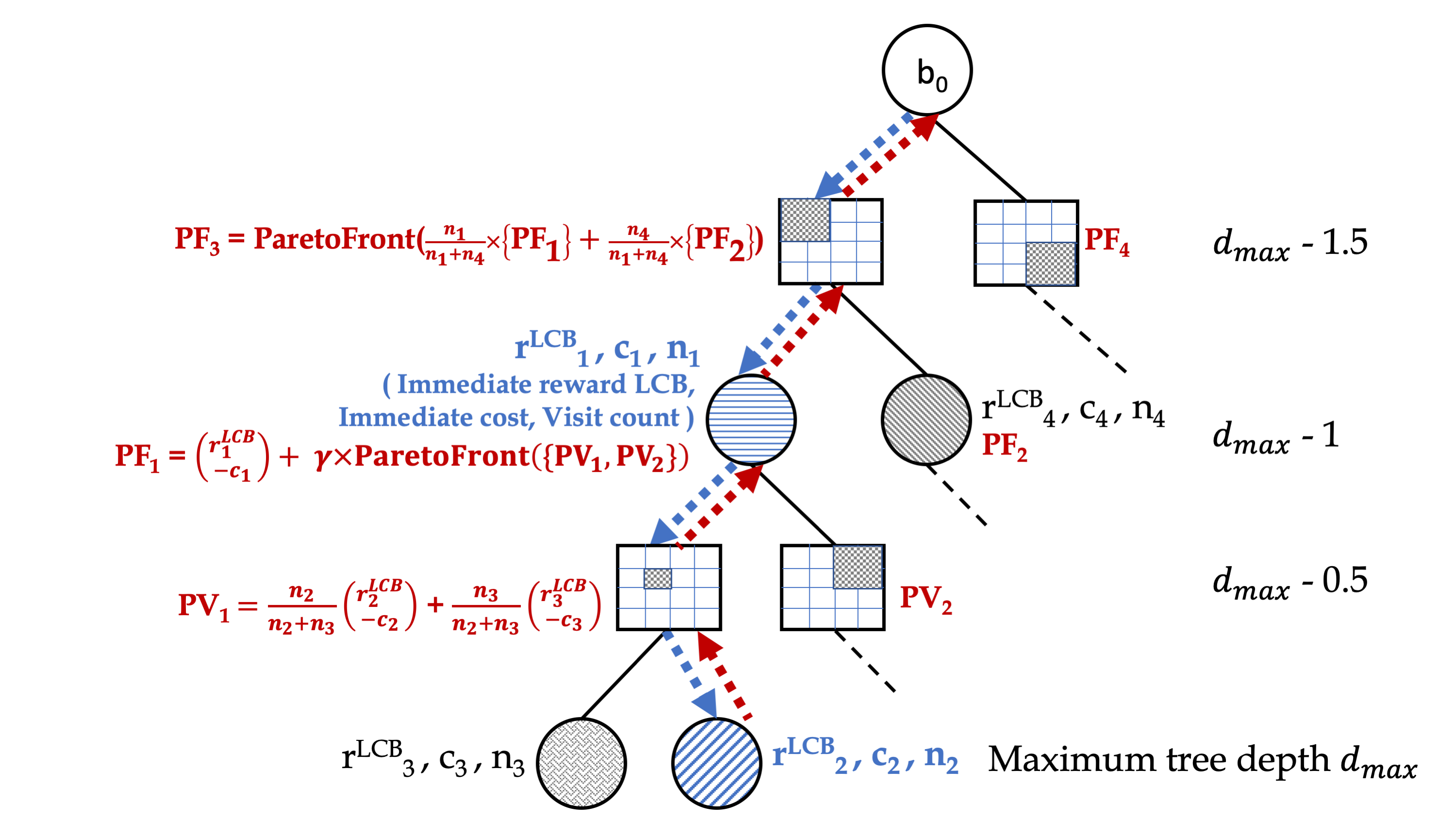}
	\caption{Illustration of a search tree $\mathcal{T}_t$ with $d_{\text{max}} = 2$. $b_0$ is the belief at the root node. The action nodes (rectangles) indicate region sensing actions. The belief nodes (circles) are shaded to indicate the evolving posterior belief in the search tree. Blue arrows illustrate top-down traversal for episode $m'$. $r^{LCB}_1$, $c_1$, $n_1$, $r^{LCB}_2$, $c_2$ and $n_2$ are updated. Red arrows illustrate bottom-up traversal for episode $m'$. $PV_{\{1,2\}}$ are vectors, $PF_{\{1,2,3,4\}}$ are the pareto fronts at the respective belief and action nodes. $\gamma$ is the discount factor. ParetoFront() obtains the pareto-optimal vectors from an input set. During backpropagation, only $PV_1$, $PF_1$ and $PF_3$ are updated corresponding to nodes encountered during top-down traversal.}
	\label{fig:tree_traversal}
\end{figure}

\textbf{Pareto front construction with confidence bounds:} During the selection and expansion phase in any episode $m'$, the one-step lookahead reward $\lambda^-_{m'}$ is computed at each visited belief node $h$ (\cref{line:35_36}). We note that $\lambda^-_{m'}$ depends on the posterior sample $\bm\beta_{m'}$ drawn for that episode. Assuming that a belief node $h$ is visited in $n(h)$ episodes so far while building the search tree $\mathcal{T}_t$, we account for the stochasticity in the computed $\lambda^-$ by maintaining the Lower Confidence Bound (LCB) of these rewards (denoted $r^{LCB}_{h}$) using the Student's t-distribution to estimate a 95\% confidence interval (\cref{line:update_rLCB}). Denoting the cost of executing the action that transitions into the belief node $h$ as $c_h$ (\cref{line:35_36}), we define a LCB based immediate (one-step lookahead) reward-cost vector at $h$, ${\bmg}_{h} = \begin{bmatrix}r^{LCB}_{h}& -c_{h}\end{bmatrix}^\mathrm{T}$ which is essential to our multi-objective decision making as described next. \cref{fig:tree_traversal} highlights, in blue, these variables updated during the selection and expansion phase in one episode. 

Next, we compute the pareto front over the multi-step lookahead reward-cost vectors at tree nodes visited during the backpropagation phase in episode $m'$. \cref{fig:tree_traversal} illustrates this process. Note that the search tree depth at the leaf nodes is $d_{\text{max}}$ and consecutive action and belief nodes differ in depth by $0.5$. The lookahead reward-cost vector at the action node at depth $d_{\text{max}}-0.5$ is the weighted average of the reward-cost vectors $\bmg_{h_\ell}$ of all its leaves $h_\ell$, weights being in proportion of their visits.
%
Next, the belief node at depth $d_{\text{max}}-1$ builds a pareto front from the lookahead reward-cost vectors of all its children action nodes. It then takes a discounted sum of its immediate reward-cost vector with this pareto front to build its lookahead reward-cost vector \emph{set} since the pareto front may comprise multiple non-dominated pareto-optimal vectors. 
Repeating these steps in episode $m'$ all the way up to the root node, we alternate between the following: 1) every action node builds its lookahead reward-cost vector set as the pareto front computed from the weighted average of the lookahead vectors of its children belief nodes 2) every belief node builds its lookahead reward-cost vector set by taking the discounted sum of its immediate reward-cost vector with the pareto front obtained from its children action nodes. Note that all reward-cost vectors use the LCB of the rewards. Therefore, at the end of $m$ episodes, each child action node of the root has an LCB based pareto front of lookahead reward-cost vectors. $\setA_t^*$ (\cref{line:20}) is the pareto front at the root node comprising the non-dominated lookahead reward-cost vectors among its children action nodes. Finally, the agent selects the action node at the root having the maximum value of reward per unit cost among its vectors in $\setA_t^*$ (\cref{line:21}). This completes the agent's decision making step at time $t$. 

\textbf{Remark 2.} \textit{In the multi-agent setting as described in \cref{sec:problem}, 
we merely have the agents asynchronously broadcast their action and observation history and each agent independently incorporates in its belief any such data it happens to receive (\cref{line:communication}). Note that when the agents share their actions and observations, they can consequently hold similar posterior beliefs. However an agent at time $t$ will not necessarily have access to all other agents' histories from the previous $t-1$ time steps (i.e. $|\mathbf{D}_t^j| \leq t-1$) since agents may plan, execute and communicate at different rates. Thus the stochastic nature of our posterior sampling based algorithm enables multi-agent cost-aware active search without a central controller.}
%

\section{Results}
\label{sec:results}
We now evaluate CAST by comparing in simulation the total cost incurred during multi-agent active search using cost-aware agents against the cost agnostic active search algorithms SPATS \cite{ghods2021decentralized} and RSI \cite{ma2017active}. SPATS is a TS based algorithm for asynchronous multi-agent active search, whereas RSI chooses sensing actions that maximize its information gain. 
We also consider sequential point sensing (PS) as a baseline for exhaustive coverage. 

In our experiments, we focus on 2-dimensional (2D) search spaces discretized into $16\times16$ square grid cells of width 10m. An agent can move horizontally or vertically at a constant speed of 5m/s. Each sensing action incurs a fixed cost of $c_s$ seconds (s), in addition to the travel time between sensing locations. We note that the cost-aware active search strategy may differ depending on the relative magnitudes of per action sensing cost and per unit travel cost. Hence, for each setting, we will vary $c_s \in \{0\text{s},50\text{s}\}$ to simulate high travel cost and high sensing cost respectively. 
Our goal is to estimate a $k$-sparse signal $\bm\beta$ by detecting all the $k$ targets with $J$ agents. 

The search vector $\bm\beta$ is generated as a randomly uniform $k$-sparse vector in the search space. The agents are unaware of $k$ and the generative prior. We set the signal to noise variance to 16. For CAST, we set $\gamma = 0.97$ and $\alpha_s = 0.5$. The hyperparameters in SPATS and RSI follow \cite{ghods2021decentralized, ma2017active}. 
We allow the agents to continue searching the space until all targets have been recovered. Then, across 10 random trials we measure the mean and standard error (s.e.) of the total cost incurred by the team in recovering all $k$ targets. We also plot the mean and s.e. of the full recovery rate achieved as a function of the total cost incurred. The full recovery rate is defined as the fraction of targets in $\bm\beta$ that are correctly identified. 
All agents start from the same location at one corner of the search space, fixed across trials. However, the exact instantiation of the search space varies across trials in terms of the position of the targets. 

\begin{figure}[!h]
	\centering
	\begin{subfigure}{0.48\linewidth}
        \includegraphics[scale=0.35]{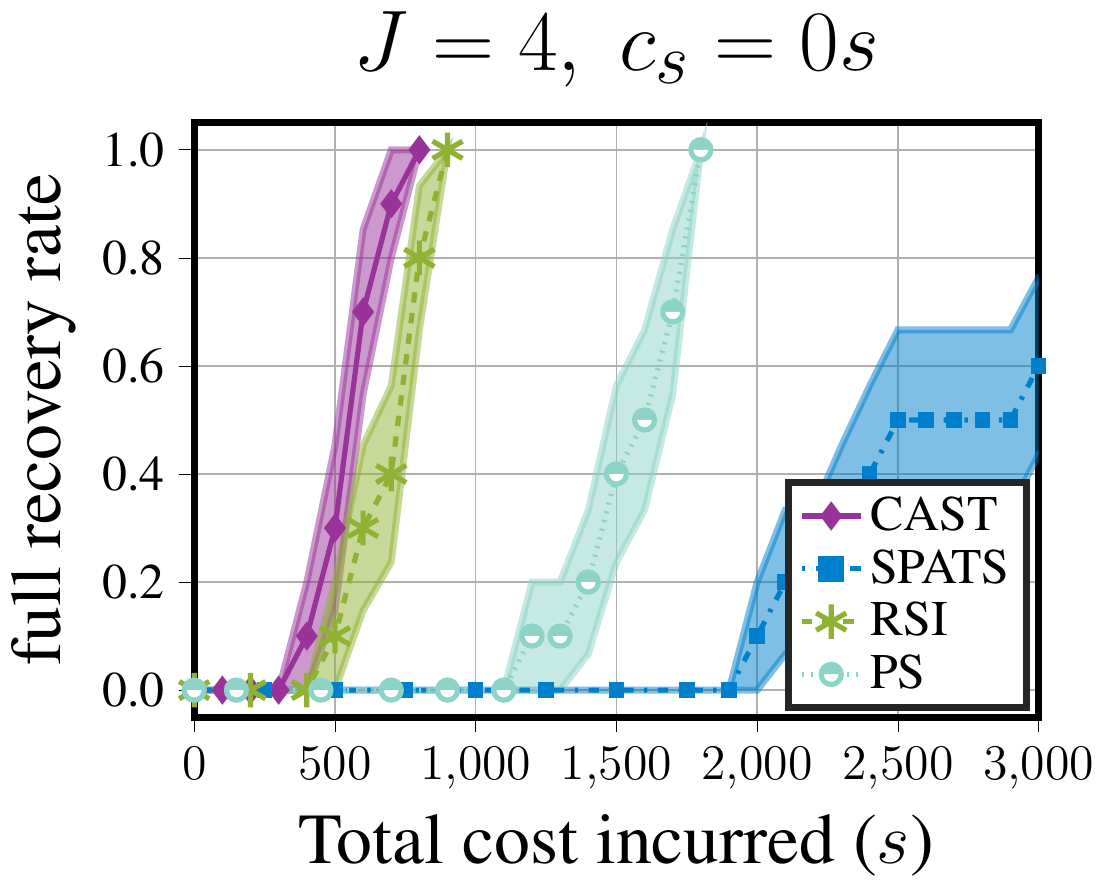}
	\end{subfigure}%
	\begin{subfigure}{0.48\linewidth}
		\includegraphics[scale=0.35]{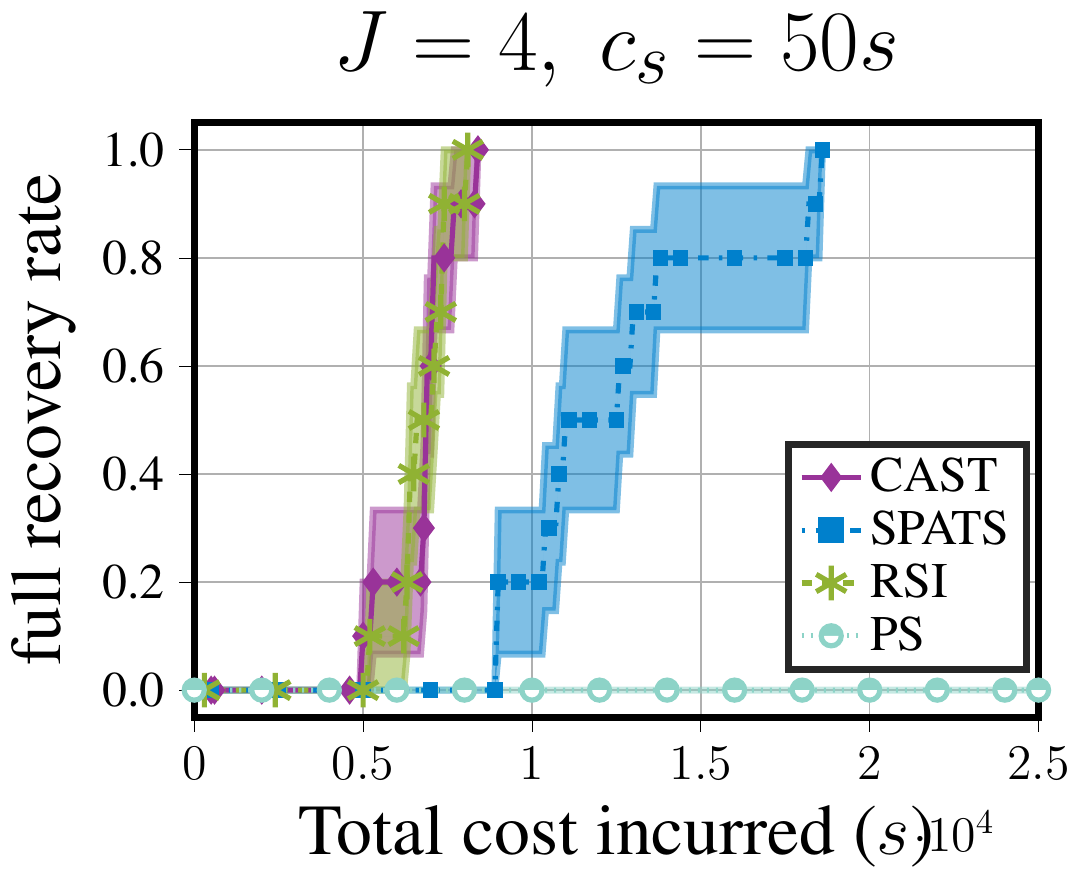}
	\end{subfigure}
	\begin{subfigure}{0.48\linewidth}
		\includegraphics[scale=0.35]{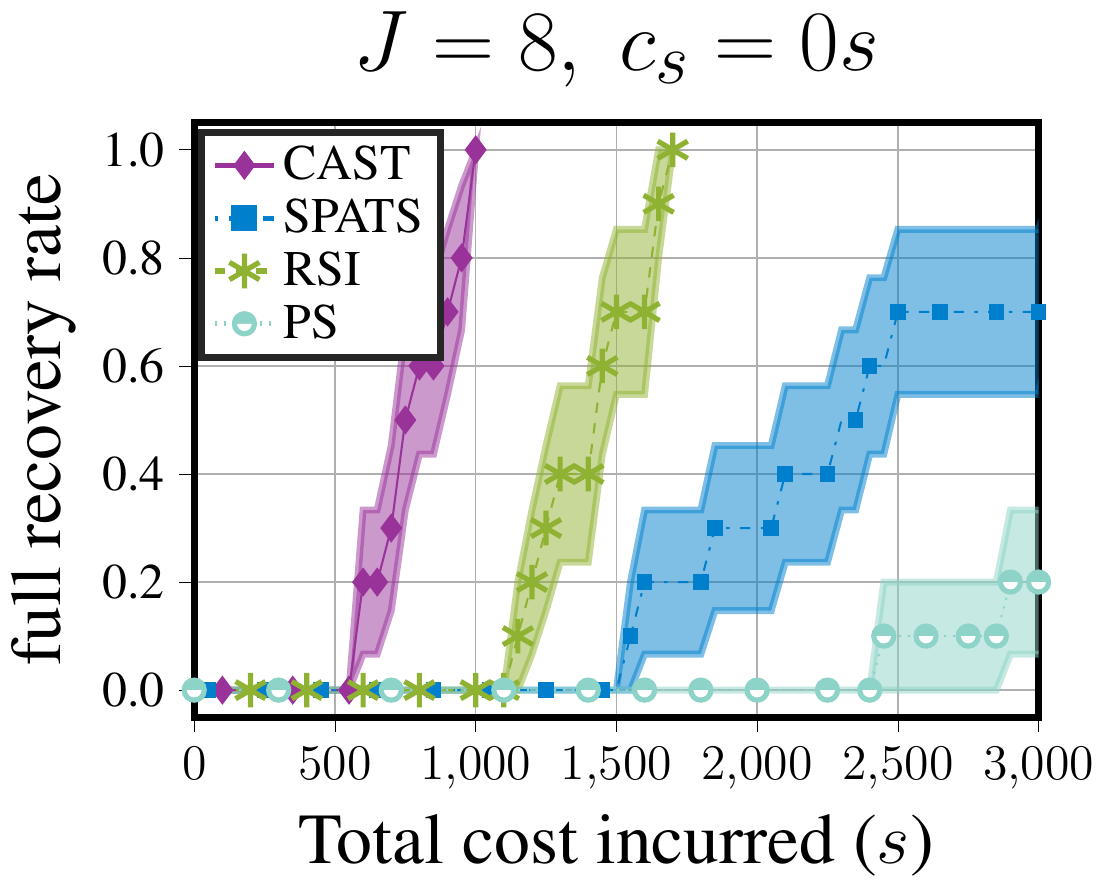}
	\end{subfigure}%
	\begin{subfigure}{0.48\linewidth}
		\includegraphics[scale=0.35]{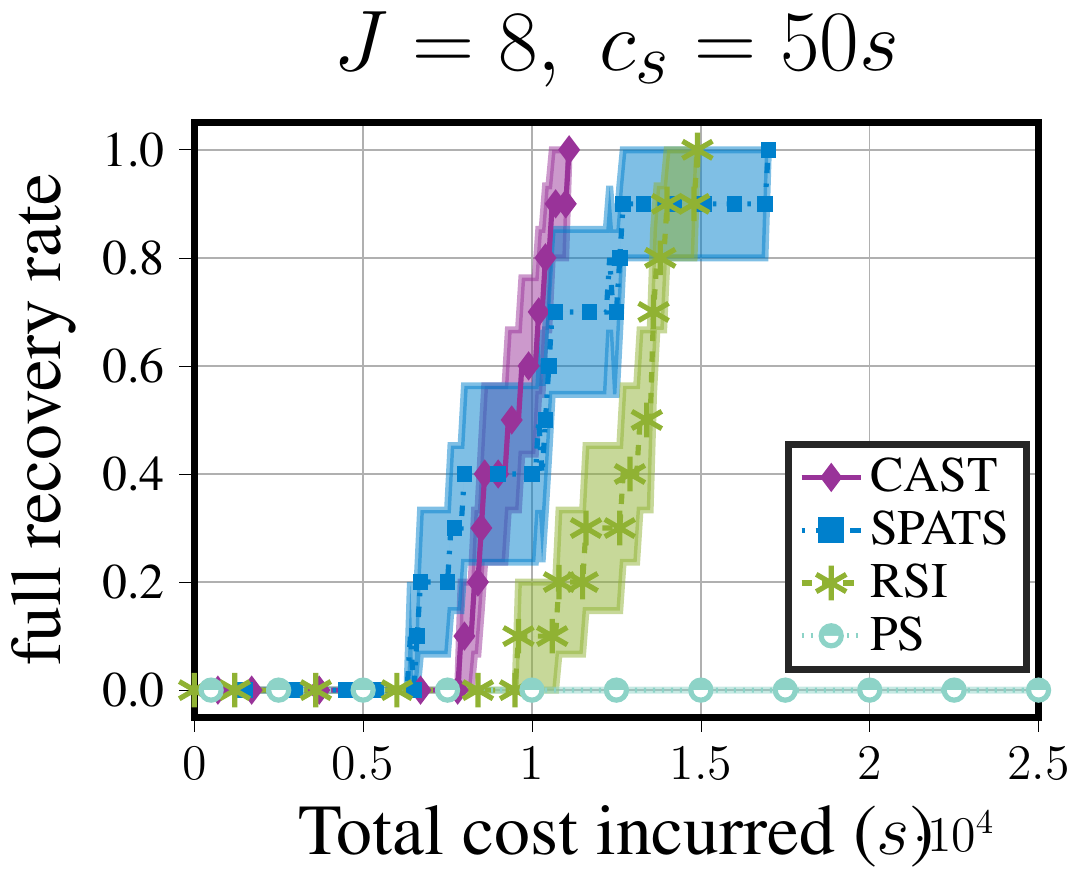}
	\end{subfigure}
	\begin{subfigure}{0.48\linewidth}
		\includegraphics[scale=0.35]{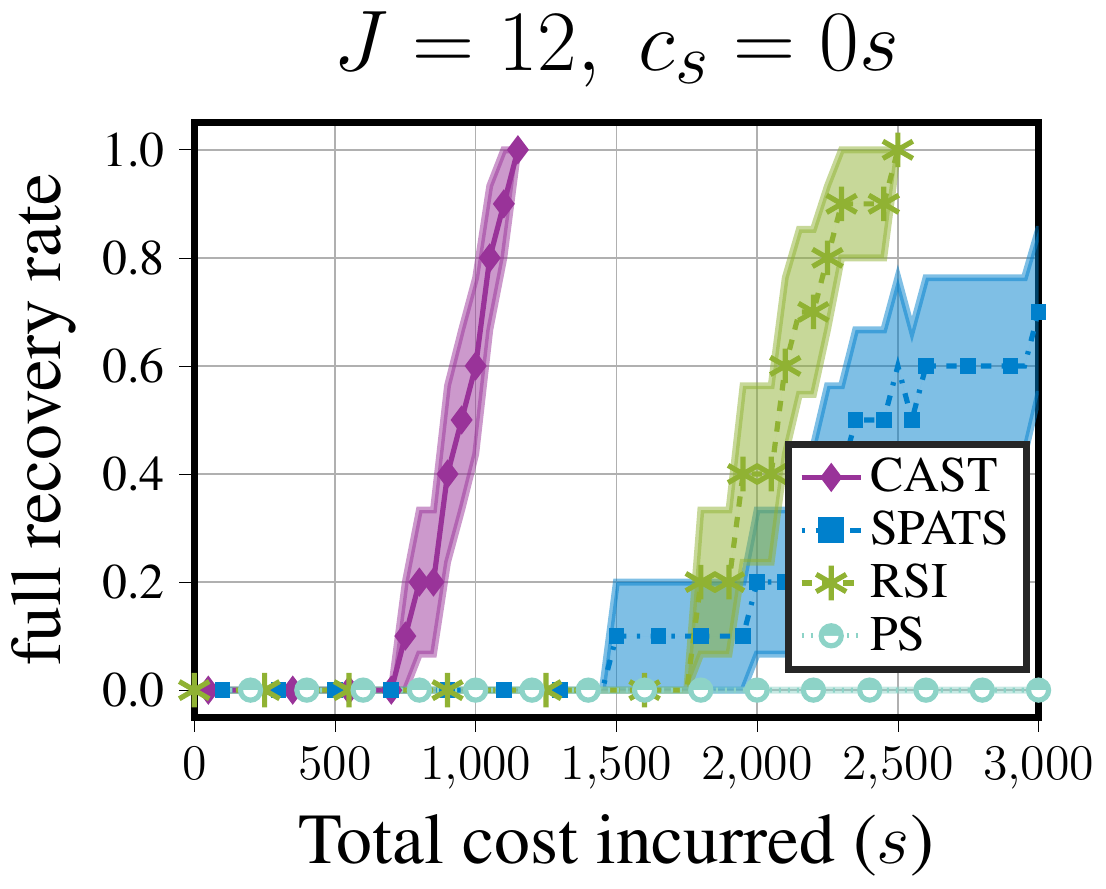}
	\end{subfigure}%
	\begin{subfigure}{0.48\linewidth}
		\includegraphics[scale=0.35]{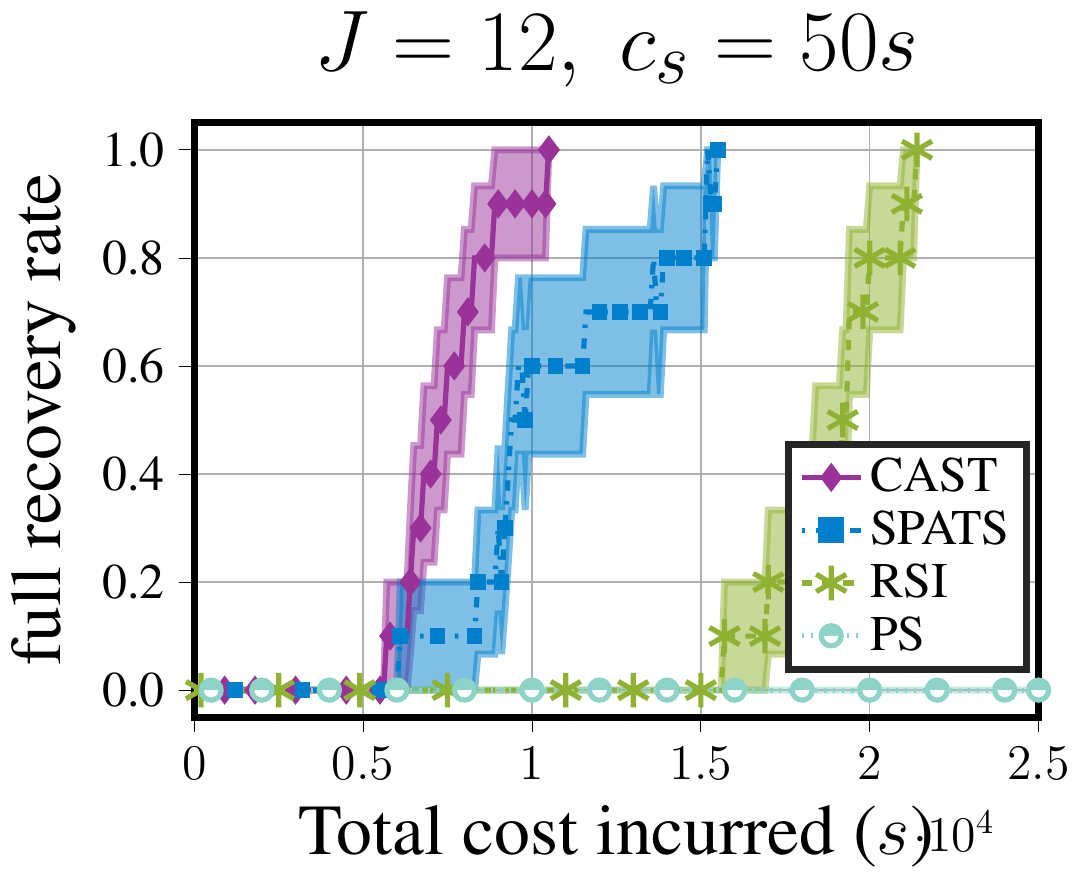}
	\end{subfigure}
	\caption{Full recovery rate versus total cost incurred in seconds in a $16\times16$ grid with $J$ agents, $k=5$ targets.}
	\label{fig:baselines2D_k5ag4n16x16}
\end{figure}

\cref{fig:baselines2D_k5ag4n16x16} shows 
full recovery rate versus total cost incurred 
with $J$ agents looking for $k = 5$ targets in a $16\times16$ search space. We vary the team size $J\in\{4,8,12\}$.
\cref{tab:16x16} indicates the corresponding 
total cost to correctly detect all targets.
CAST simulates $m=25000$ episodes with a lookahead horizon of 2 actions ($d_{\text{max}} = 2$). Each agent can choose from 341 region sensing actions over successive time steps.
\begin{table}[!h]
  \caption{Total cost (mean and s.e. over 10 trials) to achieve full recovery in a $16\times16$ grid with $J$ agents, $k = 5$ targets.}
  \label{tab:16x16}
  \centering
  \resizebox{0.85\columnwidth}{!}{%
  \begin{tabularx}{\columnwidth}{@{}l | *1{>{\centering\arraybackslash}X}@{}  *2{>{\RaggedLeft\arraybackslash}X}@{}}
    \toprule
    \centering
    Algorithm & $J$ & $c_s = 0$s & $c_s = 50$s\\
    \midrule
    CAST & 4 & \textbf{655.9 (39.4)} & \textbf{6852.3 (314.1)}\\
    SPATS & & 2988.8 (285.6) & 12563.8 (1132.7)\\
    RSI   &  & 797.4 (37.2) & 6862.4 (252.8)\\
    PS & & 1654.1 (64.4) & 42753.3 (1742.6)\\
    \midrule
    CAST & 8 & \textbf{827.0 (48.4)} & \textbf{9529.7 (350.6)}\\
    SPATS & & 2482.3 (255.7) & 10242.3 (1033.6)\\
    RSI   &  & 1455.5 (59.8) & 12815.5 (513.6)\\
    PS & & 3414.9 (143.7) & 88839.9 (3735.3)\\
    \midrule
    CAST & 12 & \textbf{991.6 (39.6)} & \textbf{7647.59 (445.4)}\\
    SPATS & & 2699.2 (240.1) & 10764.2 (948.8)\\
    RSI   &  & 2118.9 (71.0) & 19023.9 (551.1)\\
    PS & & 4827.0 (167.4) & 125582.0 (4352.2)\\
    \bottomrule
  \end{tabularx}
  }
\end{table}
We observe that CAST outperforms SPATS, RSI and PS, incurring a lower cost and a higher full recovery rate across different team sizes and cost scenarios. RSI is information greedy 
and deterministic in its decision making, so all agents choose the same actions leading to an increasing total cost with larger team sizes. On the other hand, the stochastic nature of TS based active search in SPATS is suited to the asynchronous and decentralized multi-agent setup and becomes competitive especially 
when sensing actions are more expensive than travelling ($c_s=50s$) which aligns best with the objective of active information gathering. Exhaustive coverage in PS is comparable only with a smaller team size in case when travelling is expensive ($J=4,c_s=0s$) but outperforms SPATS in that setting, showing the need for cost-awareness in active search. Unfortunately in cases that do not match their most favorable scenarios, all of these algorithms exhibit poor cost efficiency. In contrast, the cost-aware agents using CAST's posterior sampling based lookahead pareto-optimal planning and stochastic decision making are able to achieve cost efficiency across different cost scenarios with teams of varying sizes. 

We also evaluate the robustness of CAST by comparing the total cost incurred to 
correctly identify all targets as the number of targets increases in the search space. 
\begin{table}[!h]
  \caption{Total cost (mean and s.e. over 5 trials) to achieve full recovery in a $16\times16$ grid with $J = 8$ agents, $k$ targets.}
  \label{tab:cost_vs_k_g8n16x16}
  \centering
  \resizebox{0.85\columnwidth}{!}{%
    \begin{tabularx}{\columnwidth}{@{}l | *1{>{\centering\arraybackslash}X}@{}  *2{>{\RaggedLeft\arraybackslash}X}@{}}
    \toprule
    Algorithm & $k$ & $c_s = 0$s & $c_s = 50$s\\
    \midrule
    CAST & 4 & \textbf{740.7 (26.5)} & \textbf{8130.4 (293.1)}\\
    SPATS & & 3404.4 (432.9) & 13574.4 (1792.2)\\
    RSI   &  & 1262.8 (73.4) & 10242.8 (685.8)\\
    PS & & 2698.3 (438.6) & 70208.3 (11404.6)\\
    \cmidrule(r){0-3}
    CAST & 8 & \textbf{735.3 (57.9)} & \textbf{9157.0 (476.7)}\\
    SPATS & & 3217.2 (461.5) & 13267.2 (1737.3)\\
    RSI   &  & 1968.4 (72.2) & 18398.4 (500.3)\\
    PS & & 3339.9 (151.7) & 86889.9 (3945.2)\\
    \cmidrule(r){0-3}
    CAST & 16 & \textbf{843.9 (29.9)} & \textbf{8880.8 (316.2)}\\
    SPATS & & 3032.7 (54.3) & 13212.7 (321.0)\\
    RSI   &  & 2734.4 (58.9) & 30524.4 (871.3)\\
    PS & & 3559.1 (83.7) & 92589.1 (2176.8)\\
    \bottomrule
  \end{tabularx}
  }
\end{table}
\cref{tab:cost_vs_k_g8n16x16} 
shows 
that CAST not only outperforms all others across multi-target and cost scenarios, additionally the total cost incurred is hardly affected by $k$ since CAST enables cost awareness through decentralized decision making independent of team size $J$ and sparsity rate $k$. In contrast, SPATS being myopic in nature exhibits more randomness in the actions selected, whereas RSI 
approximates its mutual information objective assuming $k=1$, thereby requiring more sensing actions to recover all targets as $k$ increases.
%
For further visualization of the cost-aware multi-agent behavior of CAST compared to RSI and SPATS, we refer to the webpage we created at \href{https://sites.google.com/view/cast-multiagent/home}{this link}.
\section{Conclusion}
\label{sec:conc}
We have proposed CAST 
for detecting sparsely distributed targets without a central planner. 
Interesting directions of future work include predicting cost-aware trajectories for continuous sensing 
as well as cost-aware active search and tracking of dynamic targets. 
\clearpage
\bibliographystyle{IEEEtran}
\bibliography{IEEEabrv,ref}

\begin{thebibliography}{10}
\providecommand{\url}[1]{#1}
\csname url@rmstyle\endcsname
\providecommand{\newblock}{\relax}
\providecommand{\bibinfo}[2]{#2}
\providecommand\BIBentrySTDinterwordspacing{\spaceskip=0pt\relax}
\providecommand\BIBentryALTinterwordstretchfactor{4}
\providecommand\BIBentryALTinterwordspacing{\spaceskip=\fontdimen2\font plus
\BIBentryALTinterwordstretchfactor\fontdimen3\font minus
  \fontdimen4\font\relax}
\providecommand\BIBforeignlanguage[2]{{%
\expandafter\ifx\csname l@#1\endcsname\relax
\typeout{** WARNING: IEEEtran.bst: No hyphenation pattern has been}%
\typeout{** loaded for the language `#1'. Using the pattern for}%
\typeout{** the default language instead.}%
\else
\language=\csname l@#1\endcsname
\fi
#2}}

\bibitem{garnett2011bayesian}
R.~Garnett, Y.~Krishnamurthy, D.~Wang, J.~Schneider, and R.~Mann, ``Bayesian
  optimal active search on graphs,'' in \emph{Ninth Workshop on Mining and
  Learning with Graphs}, 2011.

\bibitem{garnett2012bayesian}
R.~Garnett, Y.~Krishnamurthy, X.~Xiong, J.~Schneider, and R.~Mann, ``Bayesian
  optimal active search and surveying,'' in \emph{Proceedings of the 29th
  International Coference on International Conference on Machine Learning},
  2012.

\bibitem{pvenivcka2019data}
R.~P{\v{e}}ni{\v{c}}ka, J.~Faigl, M.~Saska, and P.~V{\'a}{\v{n}}a, ``Data
  collection planning with non-zero sensing distance for a budget and curvature
  constrained unmanned aerial vehicle,'' \emph{Autonomous Robots}, vol.~43,
  no.~8, 2019.

\bibitem{kent2020human}
D.~Kent and S.~Chernova, ``Human-centric active perception for autonomous
  observation,'' in \emph{2020 IEEE International Conference on Robotics and
  Automation (ICRA)}, 2020.

\bibitem{yan2013survey}
Z.~Yan, N.~Jouandeau, and A.~A. Cherif, ``A survey and analysis of multi-robot
  coordination,'' \emph{International Journal of Advanced Robotic Systems},
  vol.~10, no.~12, 2013.

\bibitem{robin2016multi}
C.~Robin and S.~Lacroix, ``Multi-robot target detection and tracking: taxonomy
  and survey,'' \emph{Autonomous Robots}, vol.~40, no.~4, 2016.

\bibitem{lim2016adaptive}
Z.~W. Lim, D.~Hsu, and W.~S. Lee, ``Adaptive informative path planning in
  metric spaces,'' \emph{The International Journal of Robotics Research},
  vol.~35, no.~5, 2016.

\bibitem{lazebnik2006beyond}
S.~Lazebnik, C.~Schmid, and J.~Ponce, ``Beyond bags of features: Spatial
  pyramid matching for recognizing natural scene categories,'' in \emph{2006
  IEEE Computer Society Conference on Computer Vision and Pattern Recognition
  (CVPR'06)}, vol.~2, 2006.

\bibitem{ghods2021decentralized}
R.~Ghods, A.~Banerjee, and J.~Schneider, ``Decentralized multi-agent active
  search for sparse signals,'' in \emph{Uncertainty in Artificial
  Intelligence}, 2021.

\bibitem{popovic2017multiresolution}
M.~Popovi{\'c}, T.~Vidal-Calleja, G.~Hitz, I.~Sa, R.~Siegwart, and J.~Nieto,
  ``Multiresolution mapping and informative path planning for uav-based terrain
  monitoring,'' in \emph{2017 IEEE/RSJ International Conference on Intelligent
  Robots and Systems (IROS)}, 2017.

\bibitem{linchant2015unmanned}
J.~Linchant, J.~Lisein, J.~Semeki, P.~Lejeune, and C.~Vermeulen, ``Are unmanned
  aircraft systems (uas s) the future of wildlife monitoring? a review of
  accomplishments and challenges,'' \emph{Mammal Review}, vol.~45, no.~4, 2015.

\bibitem{gupta2017decision}
A.~Gupta, D.~Bessonov, and P.~Li, ``A decision-theoretic approach to
  detection-based target search with a uav,'' in \emph{2017 IEEE/RSJ
  International Conference on Intelligent Robots and Systems (IROS)}, 2017.

\bibitem{meera2019obstacle}
A.~A. Meera, M.~Popovi{\'c}, A.~Millane, and R.~Siegwart, ``Obstacle-aware
  adaptive informative path planning for uav-based target search,'' in
  \emph{2019 International Conference on Robotics and Automation (ICRA)}, 2019.

\bibitem{singh2009nonmyopic}
A.~Singh, A.~Krause, and W.~J. Kaiser, ``Nonmyopic adaptive informative path
  planning for multiple robots,'' in \emph{Proceedings of the 21st
  international jont conference on Artifical intelligence}, 2009.

\bibitem{choudhury2020adaptive}
S.~Choudhury, N.~Gruver, and M.~J. Kochenderfer, ``Adaptive informative path
  planning with multimodal sensing,'' in \emph{Proceedings of the International
  Conference on Automated Planning and Scheduling}, vol.~30, 2020.

\bibitem{marchant2014sequential}
R.~Marchant, F.~Ramos, S.~Sanner, \emph{et~al.}, ``Sequential bayesian
  optimisation for spatial-temporal monitoring.'' in \emph{UAI}, 2014.

\bibitem{rajan2015bayesian}
P.~Rajan, W.~Han, R.~Sznitman, P.~Frazier, and B.~Jedynak, ``Bayesian multiple
  target localization,'' \emph{Journal of Machine Learning Research}, vol.~37,
  2015.

\bibitem{jiang2017efficient}
S.~Jiang, G.~Malkomes, G.~Converse, A.~Shofner, B.~Moseley, and R.~Garnett,
  ``Efficient nonmyopic active search,'' in \emph{International Conference on
  Machine Learning}, 2017.

\bibitem{azimi2012batch}
J.~Azimi, A.~Fern, X.~Z. Fern, G.~Borradaile, and B.~Heeringa, ``Batch active
  learning via coordinated matching,'' in \emph{Proceedings of the 29th
  International Coference on International Conference on Machine Learning},
  2012.

\bibitem{jiang2018efficient}
S.~Jiang, G.~Malkomes, M.~Abbott, B.~Moseley, and R.~Garnett, ``Efficient
  nonmyopic batch active search,'' in \emph{Advances in Neural Information
  Processing Systems}, 2018.

\bibitem{ma2017active}
Y.~Ma, R.~Garnett, and J.~Schneider, ``Active search for sparse signals with
  region sensing,'' in \emph{Proceedings of the Thirty-First AAAI Conference on
  Artificial Intelligence}, 2017.

\bibitem{ghods2021multi}
R.~Ghods, W.~J. Durkin, and J.~Schneider, ``Multi-agent active search using
  realistic depth-aware noise model,'' in \emph{2021 IEEE International
  Conference on Robotics and Automation (ICRA)}, 2021.

\bibitem{jiang2019cost}
S.~Jiang, R.~Garnett, and B.~Moseley, ``Cost effective active search,'' in
  \emph{Advances in Neural Information Processing Systems}, 2019.

\bibitem{kaelbling1998planning}
L.~P. Kaelbling, M.~L. Littman, and A.~R. Cassandra, ``Planning and acting in
  partially observable stochastic domains,'' \emph{Artificial intelligence},
  vol. 101, no. 1-2, 1998.

\bibitem{kocsis2006bandit}
L.~Kocsis and C.~Szepesv{\'a}ri, ``Bandit based monte-carlo planning,'' in
  \emph{European conference on machine learning}, 2006.

\bibitem{browne2012survey}
C.~B. Browne, E.~Powley, D.~Whitehouse, S.~M. Lucas, P.~I. Cowling,
  P.~Rohlfshagen, S.~Tavener, D.~Perez, S.~Samothrakis, and S.~Colton, ``A
  survey of monte carlo tree search methods,'' \emph{IEEE Transactions on
  Computational Intelligence and AI in games}, vol.~4, no.~1, 2012.

\bibitem{silver2010monte}
D.~Silver and J.~Veness, ``Monte-carlo planning in large pomdps,'' in
  \emph{Advances in neural information processing systems}, 2010.

\bibitem{flaspohler2019information}
G.~Flaspohler, V.~Preston, A.~P. Michel, Y.~Girdhar, and N.~Roy,
  ``Information-guided robotic maximum seek-and-sample in partially observable
  continuous environments,'' \emph{IEEE Robotics and Automation Letters},
  vol.~4, no.~4, 2019.

\bibitem{fischer2020information}
J.~Fischer and {\"O}.~S. Tas, ``Information particle filter tree: An online
  algorithm for pomdps with belief-based rewards on continuous domains,'' in
  \emph{Proceedings of the 37th International Conference on Machine Learning},
  2020.

\bibitem{bernstein2002complexity}
D.~S. Bernstein, R.~Givan, N.~Immerman, and S.~Zilberstein, ``The complexity of
  decentralized control of markov decision processes,'' \emph{Mathematics of
  operations research}, vol.~27, no.~4, 2002.

\bibitem{oliehoek2016concise}
F.~A. Oliehoek, C.~Amato, \emph{et~al.}, \emph{A concise introduction to
  decentralized POMDPs}.\hskip 1em plus 0.5em minus 0.4em\relax Springer, 2016,
  vol.~1.

\bibitem{Lauri_JAAMAS2020}
M.~Lauri, J.~Pajarinen, and J.~Peters, ``Multi-agent active information
  gathering in discrete and continuous-state decentralized pomdps by policy
  graph improvement,'' \emph{Autonomous Agents and Multi-Agent Systems},
  vol.~34, no.~42, 2020.

\bibitem{lauri2020multi}
M.~Lauri and F.~Oliehoek, ``Multi-agent active perception with prediction
  rewards,'' \emph{Advances in Neural Information Processing Systems}, vol.~33,
  2020.

\bibitem{sukkar2019multi}
F.~Sukkar, G.~Best, C.~Yoo, and R.~Fitch, ``Multi-robot region-of-interest
  reconstruction with dec-mcts,'' in \emph{2019 International Conference on
  Robotics and Automation (ICRA)}, 2019.

\bibitem{best2020decentralised}
G.~Best, O.~M. Cliff, T.~Patten, R.~R. Mettu, and R.~Fitch, ``Decentralised
  monte carlo tree search for active perception,'' in \emph{Algorithmic
  Foundations of Robotics XII}.\hskip 1em plus 0.5em minus 0.4em\relax
  Springer, 2020.

\bibitem{lee2018monte}
J.~Lee, G.-H. Kim, P.~Poupart, and K.-E. Kim, ``Monte-carlo tree search for
  constrained pomdps,'' in \emph{Advances in Neural Information Processing
  Systems}, 2018.

\bibitem{wang2012multi}
W.~Wang and M.~Sebag, ``Multi-objective monte-carlo tree search,'' in
  \emph{Asian conference on machine learning}, 2012.

\bibitem{chen2019pareto}
W.~Chen and L.~Liu, ``Pareto monte carlo tree search for multi-objective
  informative planning.'' in \emph{Robotics: Science and Systems}, 2019.

\bibitem{thompson1933likelihood}
W.~R. Thompson, ``On the likelihood that one unknown probability exceeds
  another in view of the evidence of two samples,'' \emph{Biometrika}, vol.~25,
  no. 3/4, 1933.

\bibitem{gopalan2014thompson}
A.~Gopalan, S.~Mannor, and Y.~Mansour, ``Thompson sampling for complex online
  problems,'' in \emph{International Conference on Machine Learning}, 2014.

\bibitem{russo2017tutorial}
D.~Russo, B.~Van~Roy, A.~Kazerouni, I.~Osband, and Z.~Wen, ``A tutorial on
  thompson sampling,'' \emph{arXiv:1707.02038}, 2017.

\bibitem{gopalan2015thompson}
A.~Gopalan and S.~Mannor, ``Thompson sampling for learning parameterized markov
  decision processes,'' in \emph{Conference on Learning Theory}, 2015.

\bibitem{bai2014thompson}
A.~Bai, F.~Wu, Z.~Zhang, and X.~Chen, ``Thompson sampling based monte-carlo
  planning in pomdps,'' in \emph{Proceedings of the International Conference on
  Automated Planning and Scheduling}, vol.~24, no.~1, 2014.

\bibitem{leike2016thompson}
J.~Leike, T.~Lattimore, L.~Orseau, and M.~Hutter, ``Thompson sampling is
  asymptotically optimal in general environments,'' \emph{arXiv:1602.07905},
  2016.

\bibitem{kandasamy2018parallelised}
K.~Kandasamy, A.~Krishnamurthy, J.~Schneider, and B.~P{\'o}czos, ``Parallelised
  bayesian optimisation via thompson sampling,'' in \emph{International
  Conference on Artificial Intelligence and Statistics}, 2018.

\bibitem{kandasamy2019myopic}
K.~Kandasamy, W.~Neiswanger, R.~Zhang, A.~Krishnamurthy, J.~Schneider, and
  B.~Poczos, ``Myopic posterior sampling for adaptive goal oriented design of
  experiments,'' in \emph{International Conference on Machine Learning}, 2019.

\bibitem{coulom2007computing}
R.~Coulom, ``Computing “elo ratings” of move patterns in the game of go,''
  \emph{ICGA journal}, vol.~30, no.~4, 2007.

\bibitem{silver2016mastering}
D.~Silver, A.~Huang, C.~J. Maddison, A.~Guez, L.~Sifre, G.~Van Den~Driessche,
  J.~Schrittwieser, I.~Antonoglou, V.~Panneershelvam, M.~Lanctot,
  \emph{et~al.}, ``Mastering the game of go with deep neural networks and tree
  search,'' \emph{nature}, vol. 529, no. 7587, 2016.

\bibitem{silver2017mastering}
D.~Silver, J.~Schrittwieser, K.~Simonyan, I.~Antonoglou, A.~Huang, A.~Guez,
  T.~Hubert, L.~Baker, M.~Lai, A.~Bolton, \emph{et~al.}, ``Mastering the game
  of go without human knowledge,'' \emph{nature}, vol. 550, no. 7676, 2017.

\bibitem{audibert2006use}
J.-Y. Audibert, R.~Munos, and C.~Szepesvari, ``{Use of variance estimation in
  the multi-armed bandit problem},'' in \emph{NIPS Workshop on On-line Trading
  of Exploration and Exploitation}, 2006.

\bibitem{shah2020non}
D.~Shah, Q.~Xie, and Z.~Xu, ``Non-asymptotic analysis of monte carlo tree
  search,'' in \emph{Abstracts of the 2020 SIGMETRICS/Performance Joint
  International Conference on Measurement and Modeling of Computer Systems},
  2020.

\end{thebibliography}


\begin{thebibliography}{00}
\bibitem{b1} G. Eason, B. Noble, and I. N. Sneddon, ``On certain integrals of Lipschitz-Hankel type involving products of Bessel functions,'' Phil. Trans. Roy. Soc. London, vol. A247, pp. 529--551, April 1955.
\bibitem{b2} J. Clerk Maxwell, A Treatise on Electricity and Magnetism, 3rd ed., vol. 2. Oxford: Clarendon, 1892, pp.68--73.
\bibitem{b3} I. S. Jacobs and C. P. Bean, ``Fine particles, thin films and exchange anisotropy,'' in Magnetism, vol. III, G. T. Rado and H. Suhl, Eds. New York: Academic, 1963, pp. 271--350.
\bibitem{b4} K. Elissa, ``Title of paper if known,'' unpublished.
\bibitem{b5} R. Nicole, ``Title of paper with only first word capitalized,'' J. Name Stand. Abbrev., in press.
\bibitem{b6} Y. Yorozu, M. Hirano, K. Oka, and Y. Tagawa, ``Electron spectroscopy studies on magneto-optical media and plastic substrate interface,'' IEEE Transl. J. Magn. Japan, vol. 2, pp. 740--741, August 1987 [Digests 9th Annual Conf. Magnetics Japan, p. 301, 1982].
\bibitem{b7} M. Young, The Technical Writer's Handbook. Mill Valley, CA: University Science, 1989.
\end{thebibliography}


\end{document}